\documentclass[11pt]{article}

\usepackage[preprint]{acl}

\usepackage{times}
\usepackage{latexsym}

\usepackage[T1]{fontenc}
\usepackage[utf8]{inputenc}

\usepackage{microtype}

\usepackage{inconsolata}

\usepackage{xspace}
\usepackage{hyperref}
\usepackage{graphicx}
\usepackage{amssymb}
\usepackage{amsmath}
\usepackage{nccmath}
\usepackage{booktabs}
\usepackage{multirow}
\usepackage[table]{xcolor} % 允许表格上色
\usepackage{float} % 用于控制图片位置
\usepackage{cuted}
\newcommand{\boldparagraph}[1]{\noindent\textbf{#1}\ }
\definecolor{bestbg}{RGB}{198,234,212}   % 柔和的绿色（色盲友好）
\definecolor{secondbg}{RGB}{231,242,255} % 淡蓝，给次优
\definecolor{secondbg2}{RGB}{255, 218, 224}
\definecolor{mygreen}{HTML}{00AA00}
\newcommand{\best}[1]{\cellcolor{bestbg}\textbf{#1}}
\newcommand{\second}[1]{\cellcolor{secondbg}#1}
\newcommand{\secondother}[1]{\cellcolor{secondbg2}#1}
\newcommand{\maketitlesupplementary}{
  \clearpage
  \twocolumn[
    \begin{center}
      {\Large\bfseries Render-of-Thought: Rendering Textual Chain-of-Thought as Images for \\[0.3em] Visual Latent Reasoning} \\
      \vspace{1.0em}
      {\large Appendix} \\
    \end{center}
    \vspace{2em}
  ]
}

\newcommand{\huggingface}{\raisebox{-1.5pt}{\includegraphics[height=1.05em]{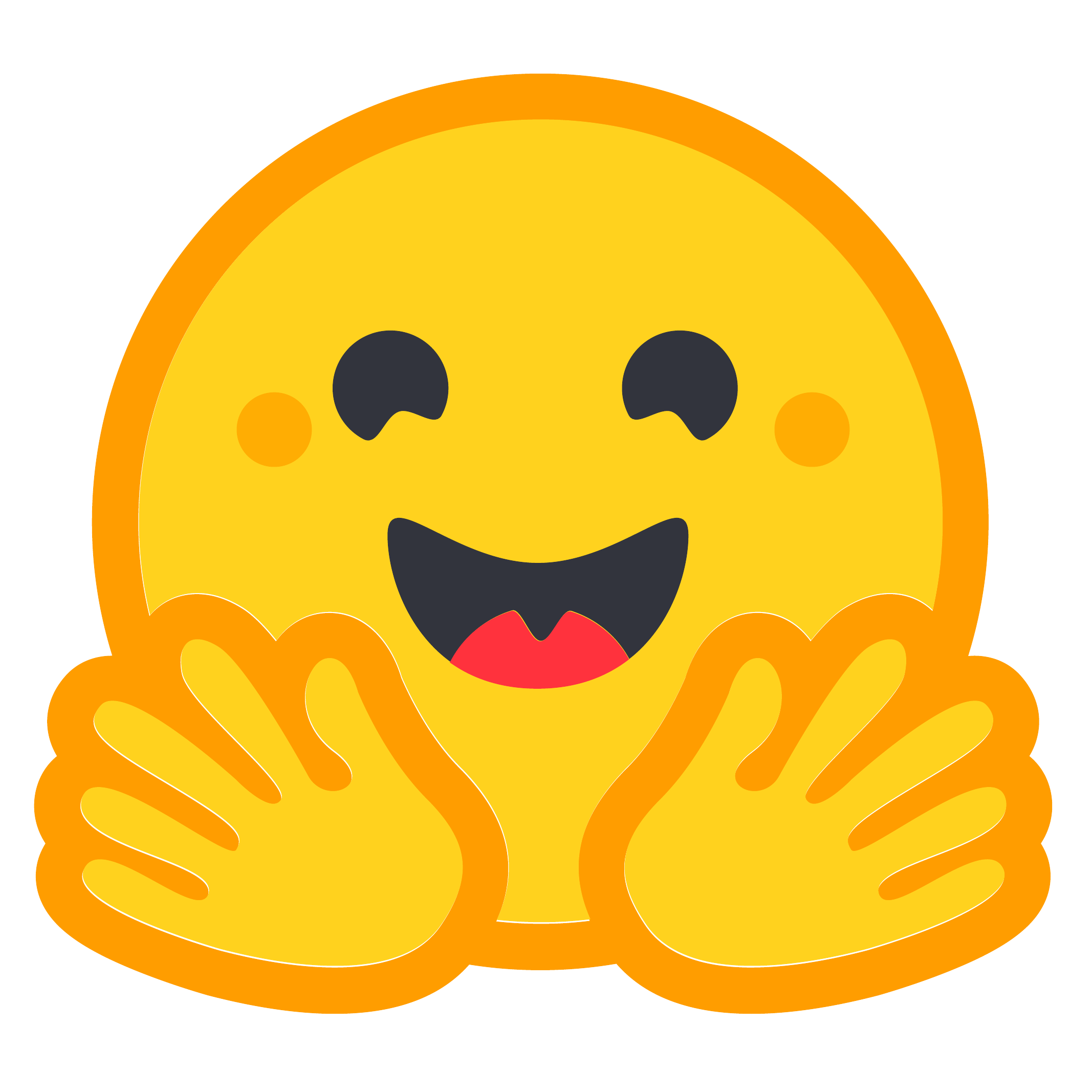}}\xspace}

\newcommand{\github}{\raisebox{-1.5pt}{\includegraphics[height=1.05em]{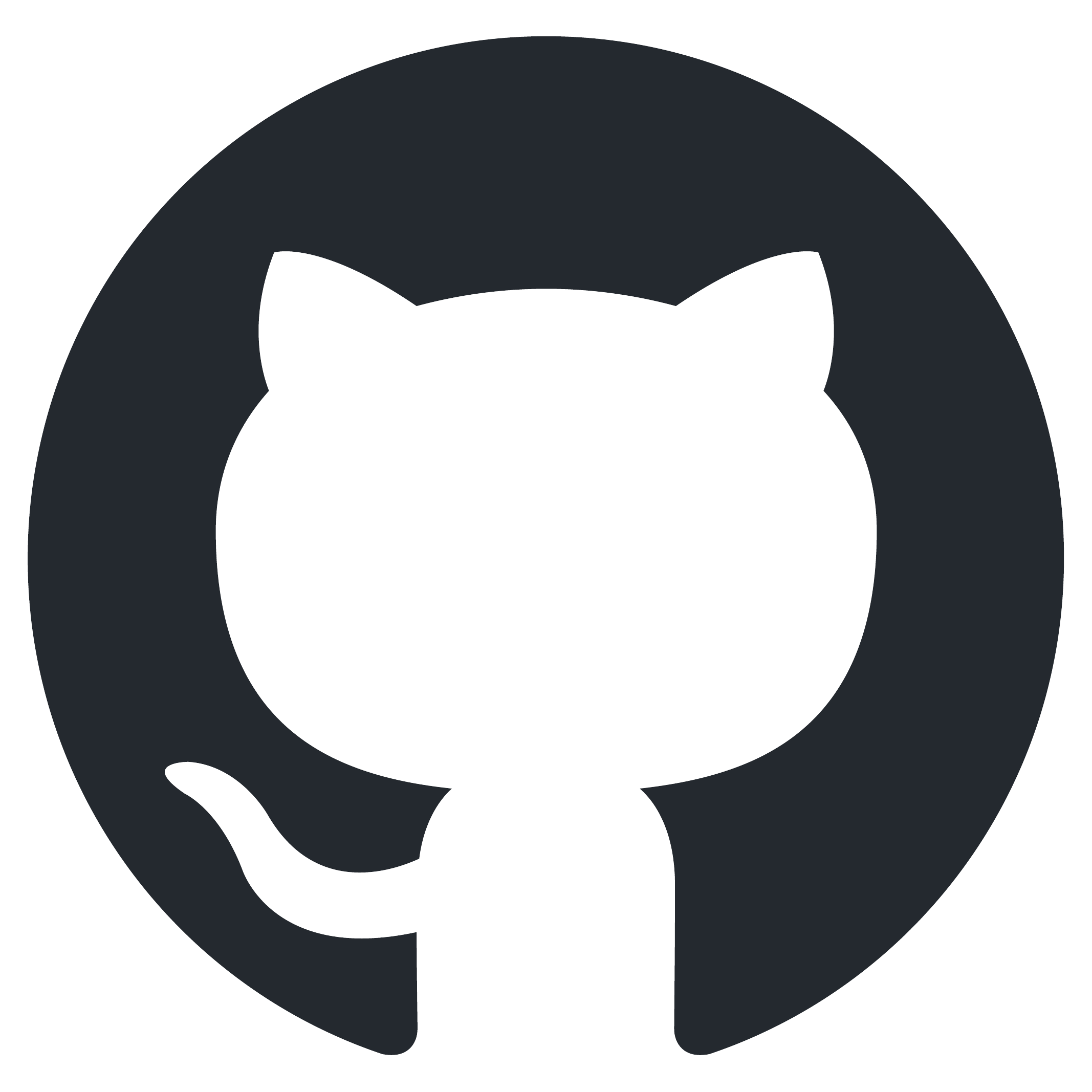}}\xspace}
\newcommand{\project}{\raisebox{0pt}{\includegraphics[height=1.0em]{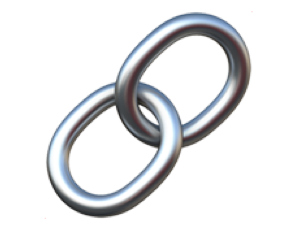}}\xspace}

\title{Render-of-Thought: Rendering Textual Chain-of-Thought as Images for \\[0.3em] Visual Latent Reasoning}

\author{
Yifan Wang$^{1,2}$ , Shiyu Li$^{1}$ , Peiming Li$^{1,3}$ , Xiaochen Yang$^{4}$ , \\
\textbf{Zheng Wei$^{1}$ \footnotemark[1]}, \textbf{Yang Tang$^{1}$\thanks{Corresponding authors.} \thanks{Project Lead.}} \\
$^1$Tencent BAC\\
$^2$Shenzhen International Graduate School, Tsinghua University \\
$^3$School of Electronic and Computer Engineering, Peking University \\
$^4$School of Mathematics and Statistics, University of Glasgow \\
\small{\texttt{{ethanntang@tencent.com, hemingwei@tencent.com}}}
\\[0.4em]
{\small
\project\!\href{https://tencentbac.github.io/RoT}{Homepage} \;\textbar\;
\github\!\href{https://github.com/TencentBAC/RoT}{Code} \;\textbar\;
\huggingface\!\href{https://huggingface.co/collections/TencentBAC/rot}{Model}
}
}

\begin{document}
\maketitle
\begin{abstract}
% The verbosity of Chain-of-Thought (CoT) reasoning significantly exacerbates the inference latency and memory footprint of Large Language Models (LLMs). While recent implicit CoT approaches attempt to compress reasoning steps into continuous latent vectors, they often suffer from optimization instability and representation collapse, as they struggle to construct a meaningful latent space tabula rasa (from scratch). In this work, we propose Render-of-Thought, a novel paradigm that reformulates reasoning as Visual Latent Generation. Drawing on the high information density of visual modalities, our framework distills verbose textual reasoning paths into compact visual embeddings. Crucially, unlike prior methods that require learning new latent structures, we leverage the pre-trained, frozen vision encoder of existing Vision-Language Models (VLMs) as a strong semantic anchor. This design not only exploits the inherent continuity and robust topology of the visual space to guide intuitive reasoning but also ensures an architecture-agnostic, plug-and-play implementation that incurs zero additional architectural overhead. By aligning LLM hidden states with visual representations of rendered thoughts, we achieve $3-4\times$ token compression and significant inference acceleration. Extensive experiments on mathematical and logical reasoning benchmarks demonstrate that Render-of-Thought maintains competitive performance against explicit CoT while unlocking the efficiency of latent reasoning within standard VLM architectures.
% cot好处
Chain-of-Thought (CoT) prompting has achieved remarkable success in unlocking the reasoning capabilities of Large Language Models (LLMs).
Although CoT prompting enhances reasoning, its verbosity imposes substantial computational overhead.
% Recent attempts to compress these steps into continuous latent vectors often fail to construct meaningful reasoning topologies from scratch.
% Recent works often focus exclusively on outcome alignment, rendering the CoT information within the generated latent reasoning embeddings uninterpretable.
% Recent works often focus exclusively on outcome alignment, thereby obscuring the interpretability of the latent reasoning chain.
Recent works often focus exclusively on outcome alignment and lack supervision on the intermediate reasoning process. These deficiencies obscure the analyzability of the latent reasoning chain.
% 拓扑改了
% To address these challenges, we introduce a novel framework Render-of-Thought that reformulates reasoning as a visual perception process.
% visualize --> reify
To address these challenges, we introduce \textbf{Render-of-Thought (RoT)}, the first framework to reify the reasoning chain by rendering textual steps into images, making the latent rationale explicit and traceable.
% For the first time, we propose the paradigm of rendering textual reasoning steps into images.
% By leveraging the pre-trained vision encoders of existing Vision Language Models (VLMs) as semantic anchors, we exploit the high information density and continuity of the visual space to guide internal reasoning.
% We exploit the high information density and continuity of the visual space to guide internal reasoning through the vision encoders of existing Vision Language Models (VLMs) as semantic anchors.
Specifically, we leverage the vision encoders of existing Vision Language Models (VLMs) as semantic anchors to align the vision embeddings with the textual space.
% 具体来讲，为了保证vision embeddding能够和文本空间对齐，我们采用了vision encoders of existing Vision Language Models (VLMs) as semantic anchors。
This design ensures \textbf{plug-and-play} implementation without incurring additional pre-training overhead.
Extensive experiments on mathematical and logical reasoning benchmarks demonstrate that our method achieves 3-4$\times$ token compression and substantial inference acceleration compared to explicit CoT. 
% Furthermore, it maintains competitive performance against other methods, validating the feasibility of this paradigm.
Furthermore, it demonstrates a competitive efficiency-accuracy Pareto exploration compared to other methods, validating the feasibility of this paradigm.
Our code is available at \url{https://github.com/TencentBAC/RoT}
\end{abstract}

\section{Introduction}

\begin{figure}[t]
   \centering
   \includegraphics[width=1.0\linewidth]{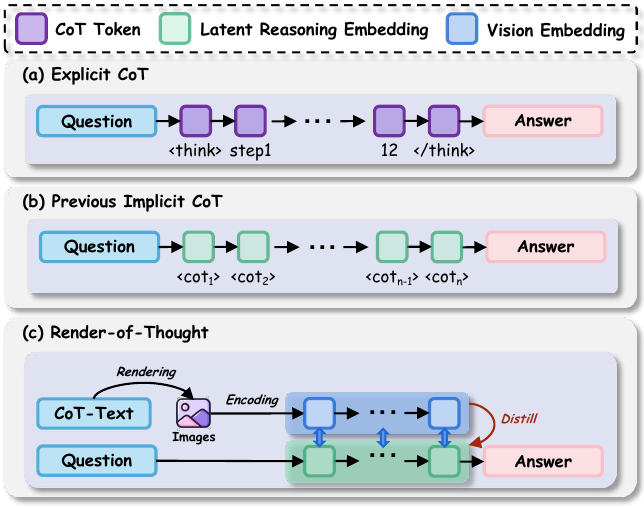}
   \caption{\textbf{Comparison of Reasoning Paradigms and Efficiency Analysis.} (a) Explicit CoT relies on verbose textual generation. (b) Implicit CoT compresses reasoning into latent space. (c) \textbf{Render-of-Thought} utilizes visual rendering as semantic anchors to structure the latent reasoning process.}
   \vspace{-16pt}
   \label{fig:intro}
\end{figure}

As Large Language Models (LLMs) continue to scale, Chain-of-Thought (CoT) prompting~\cite{cot, cot2} has become a fundamental paradigm for unlocking complex reasoning capabilities.
However, the inherent verbosity of CoT leads to prolonged inference latency and excessive memory consumption, hindering efficiency and scalability.
Recent approaches address this challenge by explicitly compressing the CoT.
Strategies range from token-level selection~\cite{tokenskip, lightthinker, han2025token} to reinforcement learning methods that incentivize shorter inference paths via rewards~\cite{aggarwal2025l1, o1-pruner, r1-compress}.
While valuable, these methods remain bound to sparse token representations.
A more promising avenue involves reasoning within dense latent spaces.
Early works such as Coconut~\cite{coconut} and CODI~\cite{codi} established the foundation for this paradigm, while CoLaR~\cite{colar} further enhanced performance through dynamic latent compression mechanisms.
However, recent efforts~\cite{yue2025hybrid, heima, latent-sft} often employ complex architectures that compromise training stability.
More critically, these methods typically focus exclusively on outcome alignment and lack supervision on the intermediate reasoning process.
By compressing thoughts into opaque vectors without explicit constraints, they obscure the analyzability of the latent reasoning chain, making it difficult to trace the model's rationale or diagnose logical errors.

% To address these challenges, we propose a paradigm shift, visualizing the reasoning chain to combine efficiency with interpretability.
% We introduce \textbf{Render-of-Thought}, a novel framework illustrated in Fig.~\ref{fig:intro}, which renders textual reasoning steps into images.
% This approach leverages the high information density of the visual modality to compress the reasoning process while keeping the rationale explicit and traceable.
% Unlike previous latent frameworks that require models to construct reasoning topologies from scratch, we utilize the frozen vision encoders of existing Vision Language Models (VLMs) as semantic anchors.
% By aligning the LLM's latent states with the visual embeddings of rendered text, we provide a structured topology that guides the reasoning process.
To address these challenges, we propose \textbf{Render-of-Thought (RoT)} (Fig.~\ref{fig:intro}), a framework that renders textual reasoning steps into images. 
This approach leverages the high information density of the visual modality to compress the reasoning process while keeping the rationale explicit. Crucially, unlike prior latent frameworks that require learning reasoning token from scratch, we utilize the frozen vision encoders of existing VLMs as semantic anchors. 
By grounding the LLM's latent states in the structured visual embeddings of rendered text, we provide a robust guide for the reasoning process.

Our pipeline implements a two-stage training strategy. Initially, we align the latent representations of the LLM with visual embeddings derived from rendered CoT. Subsequently, we enable the model to autoregressively generate the visual reasoning trajectory without requiring explicit text decoding.
This design yields two key advantages: \textit{1) Analyzability via Visualization,} which addresses the ``black box'' issue by making intermediate steps observable; and \textit{2) Plug-and-Play Efficiency,} allowing standard VLMs to be upgraded via self-distillation without extra pre-training. Experiments on Qwen3-VL-4B-Instruct~\cite{qwen3vl} show that 
% Render-of-Thought achieves a 3-4$\times$ token compression rate and marked inference acceleration compared to explicit CoT, while maintaining competitive performance. 
Render-of-Thought achieves a 3-4$\times$ token compression rate and marked inference acceleration compared to explicit CoT, demonstrating an effective efficiency-accuracy Pareto exploration.
Our contributions are summarized as follows:
% Our pipeline implements a two-stage training strategy. Initially, we align the latent representations of the LLM with visual embeddings derived from rendered CoT. Subsequently, we enable the model to autoregressively generate the visual reasoning trajectory without requiring explicit text decoding.
% This design offers two key advantages:
% \textit{1. Interpretability via Visualization.} By rendering the latent rationale into images, we make the intermediate reasoning process observable, addressing the "black box" issue of prior latent methods.
% \textit{2. Plug-and-Play Implementation.} By leveraging the native vision encoder as a teacher, our method avoids additional pre-training overhead, allowing standard open-source VLMs to be upgraded via self-distillation.
% Extensive experiments on mathematical and logical reasoning benchmarks using Qwen3-VL-4B-Instruct~\cite{qwen3vl} demonstrate that Render-of-Thought achieves a 3-4$\times$ token compression rate and significant inference acceleration compared to explicit CoT, while maintaining competitive performance against other implicit reasoning methods. Our contributions can be summarized as follows:
\begin{itemize}
    \item We introduce \textbf{Render-of-Thought}, the first framework to reify the reasoning chain by rendering textual steps into images, making latent reasoning explicit and traceable.
    \item We propose a mechanism using pre-trained vision encoders as \textbf{semantic anchors} to align vision embeddings with the textual space, ensuring a \textbf{plug-and-play} implementation without additional pre-training.
    \item Extensive experiments demonstrate that our method achieves 3-4$\times$ token compression and significant inference acceleration compared to explicit CoT, validating the feasibility and efficiency of the visual latent space as a reasoning carrier.
\end{itemize}

\section{Related Work}
\label{related work}
\boldparagraph{Explicit Chain-of-Thought Reasoning.}
Chain-of-Thought~\cite{cot} prompting has significantly enhanced the reasoning capabilities of LLMs.
However, lengthy CoT chains raise generation costs, prompting methods to compress them.
Methodologies such as~\cite{tokenskip, lightthinker, han2025token} employ heuristic or learning-based strategies to select pivotal tokens while eliminating redundant content.
Similarly, R1-Compress~\cite{r1-compress} introduces a chunk-based compression mechanism.
Other approaches~\cite{aggarwal2025l1, o1-pruner} leverage reinforcement learning to dynamically regulate reasoning length.
C3oT~\cite{c3ot} fine-tunes models using concise CoT datasets, while VeriThinker~\cite{verithinker} enables the model to autonomously determine the necessity of continued reasoning.
% Consequently, their compression potential is theoretically restricted by the information density inherent in natural language.

\boldparagraph{Implicit Chain-of-Thought Reasoning.}
Implicit Chain-of-Thought techniques accelerate inference by encoding reasoning paths in a compact latent space.
Pioneering works such as Coconut~\cite{coconut} and CODI~\cite{codi} established the foundation for continuous latent space compression.
Building on this, SoftCoT~\cite{softcot} explores the use of Soft Tokens to represent intermediate reasoning, while CoLaR~\cite{colar} investigates strategies for compressing reasoning chains within the latent space.
Furthermore, recent frameworks like~\cite{yue2025hybrid, heima, marcos} have proposed various architectural designs to support and enhance these implicit reasoning processes.
Diverging from these approaches that primarily focus on linguistic or purely latent representations, we introduce a novel paradigm by reformulating reasoning steps as autoregressive visual embedding generation.

\begin{figure*}[htbp]
   \centering
   % \vspace{10pt}
   \includegraphics[width=1.0\linewidth]{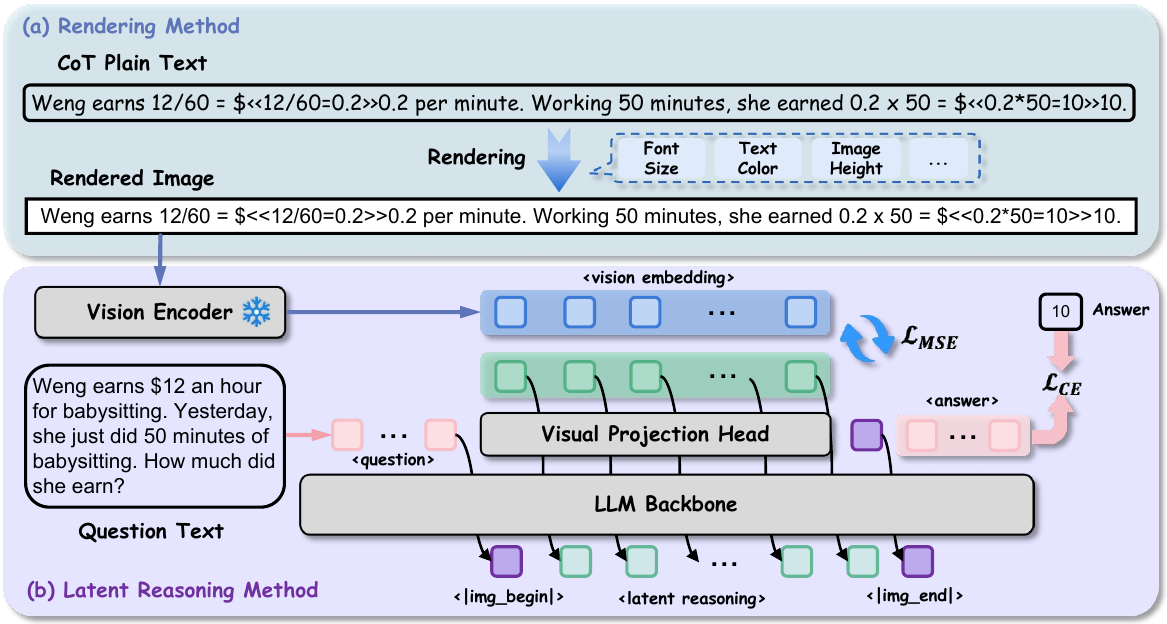}
   \caption{\textbf{Overview of the Render-of-Thought.} (a) Rendering Method transforms textual reasoning steps into compact single-line images. (b) Latent Reasoning Method aligns LLM-generated hidden states with visual features via a projection head, enabling the model to perform continuous reasoning within the visual latent space.}
   \vspace{-8pt}
   \label{fig:overview}
\end{figure*}

\boldparagraph{Text as Image for LLM.}
The paradigm of presenting textual information to LLMs via visual modalities has garnered increasing attention.
Early investigations such as PixelWorld~\cite{pixelworld} and From text to pixel~\cite{from-text-to-pixel}. have demonstrated that Vision Language Models (VLMs) possess the capability to comprehend and reason over textual content embedded within images.
More recent contributions including~\cite{text-or-pixels, glyph} have established that rendering extensive textual inputs into visual formats can significantly scale context window capacities.
However, existing ``Text-as-Image'' literature is mainly confined to input compression. To our knowledge, our framework is the first to apply visual rendering to compress the reasoning steps of VLMs.

\section{Method}
\label{method}
\subsection{Overview}
\label{overview}
Render-of-Thought introduces a novel paradigm for compressing textual CoT via optical rendering and visual knowledge distillation.
Rather than processing verbose textual steps, this approach transforms intermediate reasoning paths into compact visual representations using a pre-trained vision encoder. 
As illustrated in Fig.~\ref{fig:overview}, the framework comprises two primary components. First, textual CoT is converted into an image format using configurable rendering parameters, after which a visual encoder extracts features to serve as supervision targets. 
Second, the LLM backbone generates continuous latent reasoning tokens via the projection head, which are aligned with the visual features using Mean Squared Error (MSE) loss.
The projection head is implemented as a two-layer MLP with SwiGLU~\cite{glu} activation.
During inference, rendering and visual encoding are eliminated, requiring only a forward pass through the trained LLM Backbone and Visual Projection Head. 

\subsection{CoT Rendering}
\label{cot rendering}
The CoT rendering module transforms text into \textbf{single-line} images characterized by dynamic width and fixed height.  
This layout ensures that image patches are extracted in a strictly left-to-right manner, naturally aligning the visual sequence with the text order and eliminating spatial ambiguity. 
To accommodate varying text lengths while maintaining visual consistency, the image width is dynamically computed based on font size.
In our experiments, the default configuration employs a 32 px \colorbox[HTML]{E0E0E0}{\texttt{Image Height}}, 4 px \colorbox[HTML]{E0E0E0}{\texttt{Padding}}, and 20 px \colorbox[HTML]{E0E0E0}{\texttt{Font Size}}. 
Additionally, images are rendered with black text on a white background. Visualization examples are provided in Appendix Sec.~\ref{case study}.

\subsection{Two-Stage Training Framework}
\label{two-stage training framework}
\begin{figure*}[htbp]
   \centering
   \includegraphics[width=1.0\linewidth]{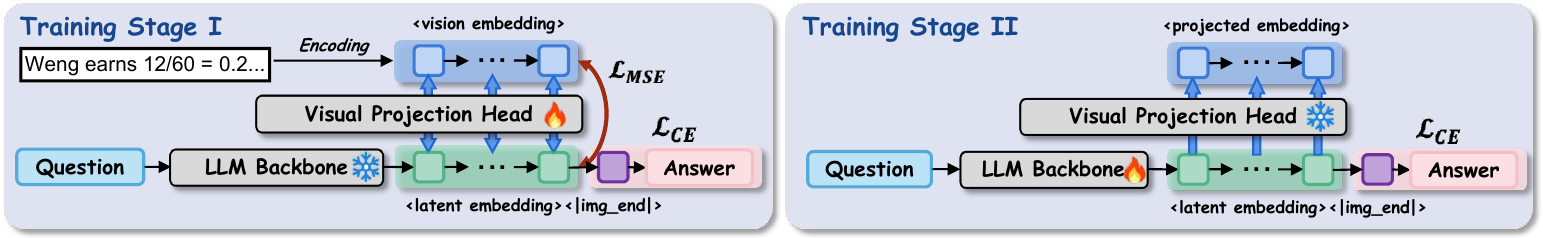}
   \caption{\textbf{Two-Stage Training Framework.} Stage I optimizes the projection head to map linguistic states to visual embeddings while freezing the backbone. Stage II fine-tunes the LLM to autoregressively generate the latent reasoning chain followed by the final answer.}
   %\vspace{-12pt}
   \label{fig:training stage}
\end{figure*}

To effectively translate the discrete reasoning capabilities of LLMs into a continuous visual latent space, we propose a progressive \textbf{two-stage} training paradigm.
As shown in Fig.~\ref{fig:training stage}, this framework is designed to first align the semantic representations between the latent hidden states and visual modalities and subsequently enable the model to perform autoregressive latent reasoning.

\subsubsection{Stage I: Visual Alignment}
\label{stage I: visual alignment}
The first stage aligns the LLM's linguistic representations with the Vision Encoder's visual embeddings.
While this alignment strategy mirrors the standard paradigm of Multimodal LLMs (MLLMs), it operates in the inverse direction. 
Unlike typical MLLMs that project visual features into the LLM's input space for understanding, we map the LLM's hidden states to the visual embedding space at the output side. 
In this phase, we freeze the parameters of both the pre-trained LLM Backbone 
$\mathcal{M}$ and the Vision Encoder $\mathcal{V}$ to preserve their inherent semantic capabilities, exclusively optimizing a lightweight Visual Projection Head 
$\phi$ to perform this text-to-vision mapping.

Given an input question $x$ and its corresponding CoT $y_{cot}$, we first render $y_{cot}$ into an image using the rendering method described in Sec.~\ref{cot rendering}. 
The Vision Encoder processes this image to extract target visual embeddings $\mathbf{V} = \{v_1, v_2, \dots, v_K\}$, where $v_i \in \mathbb{R}^{d_v}$.
The \colorbox[HTML]{E0E0E0}{$<\vert\texttt{img\_begin}\vert>$} token is appended to the question to trigger visual reasoning.
At reasoning step $t$, the latent reasoning embedding is derived as $\hat{v}_{t} = \phi(\mathcal{V}(\mathcal{M}(x, \text{\scriptsize<|img\_begin|>})))$.
The alignment loss between $\hat{v}_{t}$ and the vision embeddings is defined as:
\begin{equation}
\mathcal{L}_{align} = \frac{1}{K} \sum_{t=1}^{K} \| \hat{v}_{t} - v_t \|^2_2.
\label{alignment}
\end{equation}

Furthermore, to align the model with the proposed reasoning paradigm during Stage I, we simultaneously model the cross-entropy loss for both the \colorbox[HTML]{E0E0E0}{$<\vert\texttt{img\_end}\vert>$} special token and the answer:
\begin{equation}
\begin{gathered}
\mathcal{L}_{pred} = {-}\mathbb{E}_{(x, \hat{\mathbf{V}}, y) \sim \mathcal{D}} [ 
\log P(y_{\text{<|img\_end|>}} \mid x, \hat{\mathbf{V}}) \\
+ \sum_{j=1}^{T} \log P(y_{j} \mid x, \hat{\mathbf{V}}, y_{<j}) ],
\end{gathered}
\label{stage I ce loss}
\end{equation}
% 原公式
% \mathcal{L}_{pred} = {-} \frac{1}{1 + T} \Big[ 
% &\log P(y_{<\vert\texttt{img\_end}\vert>} \mid x, \hat{\mathbf{V}}) \\
% &+ \sum_{j=1}^{T} \log P(y_{ans,j} \mid x, \hat{\mathbf{V}}, y_{ans,<j}) \Big].
where $\hat{\mathbf{V}}$ denotes the generated latent visual tokens, $y$ represents the ground-truth of question $x$.
The overall loss for Stage I can be formulated as:
\begin{equation}
    \mathcal{L}_{\mathrm{I}}=\mathcal{L}_{pred} + \lambda \mathcal{L}_{align}.
\end{equation}

% To ensure consistency between training and inference, we employ a recursive generation strategy.
% Specifically, at time step $t$, the LLM receives the projected embedding $\hat{v}_{t-1} = \phi(h_{t-1})$ from the previous step as input. The training objective minimizes the MSE between the predicted and target visual embeddings:

\subsubsection{Stage II: Latent Supervised Fine-Tuning}
\label{stage II: latent supervised fine-tuning}
Upon establishing the alignment between modalities, the Stage II focuses on empowering the LLM to autonomously generate the visual reasoning trajectory and the subsequent final answer.
In this stage, we freeze the Vision Encoder and the now-aligned projection head $\phi$. We fine-tune the LLM backbone parameters using LoRA~\cite{lora} to adapt the model to the latent reasoning task.

The model generates a sequence of latent visual tokens $\hat{\mathbf{V}}$ followed by the special token \colorbox[HTML]{E0E0E0}{$<\vert\texttt{img\_end}\vert>$} and final textual answer $y_{ans}$.
Since the projection head is frozen, the LLM is implicitly constrained to generate hidden states that map to meaningful visual representations.
The training objective is to maximize the likelihood of the correct answer and the special token, conditioned on the generated latent reasoning path.
The training loss $\mathcal{L}_{\mathrm{II}}$ in Stage II follows the same formulation as Eqn.~\ref{stage I ce loss}.
% \begin{equation}
% \begin{split}
% \mathcal{L}_{\mathrm{II}} = & -\mathbb{E}_{(x, \hat{\mathbf{V}}, y_{\text{<|img\_end|>}}, y) \sim \mathcal{D}} [ \\
%  & \quad \log P(y_{\text{<|img\_end|>}} \mid x, \hat{\mathbf{V}}) + \sum_{j=1}^{T} \\
%  & \quad \log P(y_{j} \mid x, \hat{\mathbf{V}}, y_{\text{<|img\_end|>}}, y_{<j}) ].
% \end{split}
% \label{stage II loss}
% \end{equation}
% 原公式：
% \mathcal{L}_{<\vert\texttt{img\_end}\vert>} = {-} \log P(y_{<\vert\texttt{img\_end}\vert>} \mid x, \hat{\mathbf{V}}), \\[0.2em]
% \begin{aligned}
% \mathcal{L}_{ans} = {-} \frac{1}{|y_{ans}|} \sum_{j=1}^{|y_{ans}|} &\log P(y_{ans, j} \mid x, \hat{\mathbf{V}}, \\[-0.5em]
% &\quad y_{<\vert\texttt{img\_end}\vert>}, y_{ans,<j}),
% \end{aligned} \\[0.2em]
% \mathcal{L}_{\mathrm{II}} = \frac{1}{1 + |y_{ans}|}(\mathcal{L}_{<\vert\texttt{img\_end}\vert>} + |y_{ans}| \mathcal{L}_{ans}).

Unlike the multi-task learning scheme in Sec.~\ref{stage I: visual alignment}, we do not enforce an explicit visual regression loss in this stage.
This allows the model to refine its internal reasoning process within the constraints of the aligned latent space, optimizing purely for the accuracy of the answer generation.

% \begin{figure*}
%    \centering
%    \includegraphics[width=1.0\linewidth]{fig/training_stage.pdf}
%    \caption{\textbf{Two-Stage Training Framework.} Stage I optimizes the projection head to map linguistic states to visual embeddings while freezing the backbone. Stage II fine-tunes the LLM to autoregressively generate the latent reasoning chain followed by the final answer.}
%    %\vspace{-12pt}
%    \label{fig:training stage}
% \end{figure*}

\begin{table*}[t]
\centering
\fontsize{10pt}{13pt}\selectfont
\resizebox{\linewidth}{!}{%
\begin{tabular}{l|cc|cc|cc|cc|ccc}
\toprule[0.15em]
        & \multicolumn{2}{c|}{GSM8k-Aug} & \multicolumn{2}{c|}{GSM-Hard} & \multicolumn{2}{c|}{SVAMP} & \multicolumn{2}{c|}{MultiArith} & \multicolumn{3}{c}{Average} \\
        & Pass@1          & \#~L          & Pass@1          & \#~L           & Pass@1         & \#~L         & Pass@1           & \#~L           & Pass@1           & \#~L &Pass@1/\#~L\\ 
\toprule[0.1em]
\multicolumn{12}{c}{\textbf{\second{\textit{Qwen3-VL-2B-Instruct}}}} \\
\noalign{\vspace{2pt}}
SFT-w/o CoT & 15.6$_{\pm.31}$ & 0.00$_{\pm.00}$ & 4.70$_{\pm.22}$ & 0.00$_{\pm.00}$ & 52.3$_{\pm.34}$ & 0.00$_{\pm.00}$ & 41.7$_{\pm.28}$ & 0.00$_{\pm.00}$ & 28.6 & 0.00 & - \\
SFT-CoT & 59.7$_{\pm.35}$ & 131.4$_{\pm1.6}$ & 33.1$_{\pm.30}$ & 207.2$_{\pm1.7}$ & 67.3$_{\pm.27}$ & 63.4$_{\pm.83}$ & 95.0$_{\pm.36}$ & 68.0$_{\pm.73}$ & 63.8 & 117.5 & 0.54 \\
\textbf{RoT (Ours)} & 23.3$_{\pm.33}$ & 32.0$_{\pm.00}$ & 8.64$_{\pm.22}$ & 32.0$_{\pm.00}$ & 53.7$_{\pm.36}$ & 32.0$_{\pm.00}$ & 62.2$_{\pm.35}$ & 32.0$_{\pm.00}$ & 37.0 & 32.0 & \best{1.16} \\
\toprule[0.1em]
\multicolumn{12}{c}{\textbf{\second{\textit{Qwen3-VL-4B-Instruct}}}} \\
\noalign{\vspace{2pt}}
SFT-w/o CoT & 26.2$_{\pm.25}$ & 0.00$_{\pm.00}$ & 9.48$_{\pm.13}$ & 0.00$_{\pm.00}$ & 70.0$_{\pm.31}$ & 0.00$_{\pm.00}$ & 85.6$_{\pm.39}$ & 0.00$_{\pm.00}$ & 47.8 & 0.00 & - \\
SFT-CoT & 81.2$_{\pm.35}$ & 127.3$_{\pm.96}$ & 53.4$_{\pm.34}$ & 191.1$_{\pm1.6}$ & 84.3$_{\pm.34}$ & 55.9$_{\pm.62}$ & 98.3$_{\pm.35}$ & 59.1$_{\pm.63}$ & 79.3 & 108.4 & 0.73 \\ 
\textbf{RoT (Ours)} & 37.8$_{\pm.30}$ & 32.0$_{\pm.00}$ & 14.1$_{\pm.15}$ & 32.0$_{\pm.00}$ & 72.7$_{\pm.44}$ & 32.0$_{\pm.00}$ & 97.2$_{\pm.43}$ & 32.0$_{\pm.00}$ & 55.4 & 32.0 & \best{1.73} \\
\toprule[0.1em]
\multicolumn{12}{c}{\textbf{\second{\textit{LLaVa-V1.6-Mistral-7B}}}} \\
\noalign{\vspace{2pt}}
SFT-w/o CoT & 10.8$_{\pm.20}$ & 0.00$_{\pm.00}$ & 2.27$_{\pm.11}$ & 0.00$_{\pm.00}$ & 40.7$_{\pm.32}$ & 0.00$_{\pm.00}$ & 52.8$_{\pm.45}$ & 0.00$_{\pm.00}$ & 26.6 & 0.00 & - \\
SFT-CoT & 38.6$_{\pm.26}$ & 151.3$_{\pm1.6}$ & 12.2$_{\pm.11}$ & 209.7$_{\pm1.8}$ & 56.3$_{\pm.45}$ & 79.9$_{\pm.85}$ & 86.1$_{\pm.52}$ & 82.3$_{\pm.75}$ & 48.3 & 130.8 & 0.37 \\
\textbf{RoT (Ours)} & 16.3$_{\pm.22}$ & 32.0$_{\pm.00}$ & 3.64$_{\pm.12}$ & 32.0$_{\pm.00}$ & 49.0$_{\pm.38}$ & 32.0$_{\pm.00}$ & 68.3$_{\pm.48}$ & 32.0$_{\pm.00}$ & 34.3 & 32.0 & \best{1.07} \\
\bottomrule[0.15em]
\end{tabular}
}
\caption{Experimental results on four grade-school reasoning datasets across three VLM architectures. Render-of-Thought achieves significant token compression compared to explicit CoT while maintaining competitive accuracy. The Pass@1/\#~L ratio measures efficiency, with higher values indicating better accuracy-to-token trade-offs.}
% \caption{Experimental results on four low-difficulty datasets. The number in brackets represents the compression ratio $r$. “\textcolor{bestbg}{\rule{5mm}{2mm}}” marks the best result, and “\textcolor{secondbg}{\rule{5mm}{2mm}}” marks the second best. \colorbox[HTML]{E0E0E0}{~~-~Hidden State} refers to removing the Representation Alignment step and directly using the last-layer hidden state as the representation of latent tokens. \colorbox[HTML]{E0E0E0}{~~-~w/o LTIM}  and \colorbox[HTML]{E0E0E0}{~~-~w/o LTSuM} denote removing the Latent Token Induction Mask and Latent Token Supervision Mask, respectively, reducing the model to a standard autoregressive attention mechanism.}
\vspace{-6pt}
\label{tab_low_difficulty}
\end{table*}

\subsection{Inference and Decoding Strategies}
The inference process requires the model to autonomously navigate the transition from the continuous latent reasoning space to the discrete textual solution space.
% As shown in Fig. \ref{fig:decoding}, we investigate two distinct decoding strategies to manage this modal shift.
We investigate two distinct decoding strategies to manage this modal shift.

\boldparagraph{Dynamic Termination via Special Tokens.}
This decoding mechanism relies on the intrinsic capability of the model to self-regulate the duration of its reasoning process.
The reasoning phase concludes at the first time step $T_{end}$ where the termination token achieves the highest probability:
\begin{equation}
\begin{split}
T_{end} = \min \{t \mid \operatorname*{arg\,max}_{w \in \mathcal{T}} & P(w | h_t) = w_{<\vert\texttt{img\_end}\vert>} \},
\end{split}
\end{equation}
where $\mathcal{T}$ denotes the token set and $h_t$ represents the hidden state at time step $t$.
The model initiates the decoding of the textual answer starting from the subsequent state $h_{T_{end}+1}$.

% \begin{figure}
%     \centering
%     \includegraphics[width=0.90\linewidth]{fig/decoding_strategies.pdf}%
%     \caption{\textbf{Inference Decoding Strategies.} (a) Dynamic Termination relies on the model generating a special token to conclude the latent phase. (b) Static Termination enforces a fixed latent token budget, ensuring a stable transition from visual reasoning to text decoding.}
%     \label{fig:decoding}
%     % \vspace{-10pt}
%  \end{figure}

\boldparagraph{Static Termination via Fixed Token Budgets.}
Despite the theoretical appeal of dynamic termination, empirical evidence suggests that self-regulated stopping can exhibit instability during the inference of continuous latent representations~\cite{lvr}.
To mitigate this, we constrain the latent chain of thought length to a fixed hyperparameter.
Upon reaching this threshold, the \colorbox[HTML]{E0E0E0}{$<\vert\texttt{img\_end}\vert>$} token is manually appended to trigger the transition from latent reasoning to text generation.
We observe that decoding strategy selection significantly influences reasoning stability, as detailed in Sec.~\ref{decoding strategies}.

\section{Experiments}
\label{experiments}
\subsection{Experiment Settings}
\label{experiment settings}
\boldparagraph{Datasets and Tasks.}
Our method is primarily trained and evaluated on GSM8k-Aug-NL~\cite{deng2023implicit}, an augmented version of the GSM8k~\cite{gsm8k} dataset containing approximately 385k training samples and over 1k test samples.
We also assess the robustness of our model on three Out-of-Distribution (OOD) datasets: (1) GSM-Hard~\cite{gsm8k-hard}, which is a difficult variant of GSM8k with more than 1k test samples, as well as (2) SVAMP~\cite{svamp} and (3) MultiArith~\cite{multiarith}, which are simpler reasoning datasets.
Additionally, we extend our experiments to the challenging MATH~\cite{math-500} dataset, which encompasses diverse disciplines including algebra, calculus, statistics, geometry, linear algebra, and number theory, utilizing 7.5k training and 0.5k test samples.
Our evaluation framework simultaneously measures accuracy (Pass@1) and computational efficiency (\char`\#\ L, denoting the average token length of the reasoning chain).
All experiments are performed across five distinct random seeds, and we report the mean values for Pass@1 and \char`\#\ L alongside their 95\% confidence intervals (CI).

\boldparagraph{Implementation Details.}
(1) Base Model: Unless otherwise specified, we utilize the pre-trained and frozen Qwen3-VL-2B/4B-Instruct~\cite{qwen3vl} and LLaVa-V1.6-Mistral-7B~\cite{llava} as our base models, incorporating LoRA modules~\cite{lora} for efficient fine-tuning. 
The Visual Projection Head consists of a two-layer MLP based on the SwiGLU~\cite{glu} activation function. 
For the Vision Encoder, we directly employ the native module from Qwen3-VL and keep it frozen. This strategy ensures alignment between the vision embeddings and the LLM backbone without the need for re-pre-training.
(2) Training Epoch: All models undergo training for 3 epochs to ensure fair comparison.
(3) Hyperparameter: Throughout Stage I and Stage II, we use the AdamW~\cite{adamw} optimizer with a fixed learning rate of 2e-5, a weight decay of 1e-2, and a batch size of 16. 
Specifically, Stage I involves 1 epoch of training, and Stage II involves 2 epochs. 
For inference, the temperature is set to 1.0 and top-p to 0.9. Please refer to Appendix Sec.~\ref{more implementation details} for additional implementation details.

\begin{table*}[!ht]
\centering
\fontsize{10pt}{11pt}\selectfont
\resizebox{\linewidth}{!}{%
\begin{tabular}{l|cc|cc|cc|cc|cc}
\toprule[0.15em]
        & \multicolumn{2}{c|}{GSM8k-Aug} & \multicolumn{2}{c|}{GSM-Hard} & \multicolumn{2}{c|}{SVAMP} & \multicolumn{2}{c|}{MultiArith} & \multicolumn{2}{c}{Average} \\
        & Pass@1          & \#~L          & Pass@1          & \#~L           & Pass@1         & \#~L         & Pass@1           & \#~L           & Pass@1          & \#~L    \\ 
\toprule[0.1em]
\multicolumn{11}{c}{\textbf{\second{\textit{LLM Based: Qwen3-4B-Instruct}}}} \\
\noalign{\vspace{2pt}}
iCoT & 13.5$_{\pm.21}$ & 0.00$_{\pm.00}$ & 4.09$_{\pm.18}$ & 0.00$_{\pm.00}$ & 36.9$_{\pm.23}$ & 0.00$_{\pm.00}$ & 49.2$_{\pm.67}$ & 0.00$_{\pm.00}$ & 25.9 & 0.00 \\
Coconut & 16.9$_{\pm.26}$ & 6.00$_{\pm.00}$ & 5.42$_{\pm.28}$ & 6.00$_{\pm.00}$ & 43.6$_{\pm.53}$ & 6.00$_{\pm.00}$ & 60.3$_{\pm.65}$ & 6.00$_{\pm.00}$ & 31.6 & 6.00 \\
CODI & 7.28$_{\pm.46}$ & 6.00$_{\pm.00}$ & 2.20$_{\pm.22}$ & 6.00$_{\pm.00}$ & 11.0$_{\pm.63}$ & 6.00$_{\pm.00}$ & 18.3$_{\pm.75}$ & 6.00$_{\pm.00}$ & 9.70 & 6.00 \\
CoLaR-2 & \best{40.0$_{\pm.19}$} & 39.6$_{\pm.12}$ & \secondother{9.17$_{\pm.05}$} & 47.4$_{\pm.15}$ & \secondother{57.7$_{\pm.23}$} & 19.2$_{\pm.06}$ & \secondother{82.2$_{\pm.11}$} & 21.1$_{\pm.08}$ & \secondother{47.3} & 31.8 \\
CoLaR-5 & 18.6$_{\pm.13}$ & 16.4$_{\pm.08}$ & 5.69$_{\pm.05}$ & 22.8$_{\pm.10}$ & 48.0$_{\pm.42}$ & 7.46$_{\pm.03}$ & 63.3$_{\pm.35}$ & 7.73$_{\pm.01}$ & 33.9 & 13.6 \\
\toprule[0.1em]
\multicolumn{11}{c}{\textbf{\second{\textit{VLM Based: Qwen3-VL-4B-Instruct}}}} \\
\noalign{\vspace{2pt}}
\textbf{RoT (Ours)} & \secondother{37.8$_{\pm.30}$} & 32.0$_{\pm.00}$ & \best{14.1$_{\pm.15}$} & 32.0$_{\pm.00}$ & \best{72.7$_{\pm.44}$} & 32.0$_{\pm.00}$ & \best{97.2$_{\pm.43}$} & 32.0$_{\pm.00}$ & \best{55.4} & 32.0 \\
\rowcolor[HTML]{E0E0E0}
~~-~w/o Stage I & 24.8$_{\pm.28}$ & 32.0$_{\pm.00}$ & 7.20$_{\pm.12}$ & 32.0$_{\pm.00}$ & 58.3$_{\pm.38}$ & 32.0$_{\pm.00}$ & 78.5$_{\pm.41}$ & 32.0$_{\pm.00}$ & 42.2 & 32.0 \\
\rowcolor[HTML]{E0E0E0}
~~-~w/o Stage II & 29.9$_{\pm.28}$ & 32.0$_{\pm.00}$ & 9.48$_{\pm.12}$ & 32.0$_{\pm.00}$ & 63.7$_{\pm.41}$ & 32.0$_{\pm.00}$ & 87.8$_{\pm.39}$ & 32.0$_{\pm.00}$ & 47.7 & 32.0 \\
\bottomrule[0.15em]
\end{tabular}
}
\caption{Comparison with LLM based latent reasoning methods on four grade-school reasoning datasets. All LLM based baselines use Qwen3-4B-Instruct as the base model. Render-of-Thought employs Qwen3-VL-4B-Instruct. Best and second-best results are highlighted with ``\textcolor{bestbg}{\rule{6mm}{2mm}}'' and ``\textcolor{secondbg2}{\rule{6mm}{2mm}}'', respectively.}
% \caption{Experimental results on four low-difficulty datasets. The number in brackets represents the compression ratio $r$. “\textcolor{bestbg}{\rule{5mm}{2mm}}” marks the best result, and “\textcolor{secondbg}{\rule{5mm}{2mm}}” marks the second best. \colorbox[HTML]{E0E0E0}{~~-~Hidden State} refers to removing the Representation Alignment step and directly using the last-layer hidden state as the representation of latent tokens. \colorbox[HTML]{E0E0E0}{~~-~w/o LTIM}  and \colorbox[HTML]{E0E0E0}{~~-~w/o LTSuM} denote removing the Latent Token Induction Mask and Latent Token Supervision Mask, respectively, reducing the model to a standard autoregressive attention mechanism.}
\label{tab_compared_with_LLM_methods}
\end{table*}
\subsection{Main Results}
\label{main results}
\boldparagraph{Performance on Low-Difficulty Tasks.}
Tab.~\ref{tab_low_difficulty} presents comprehensive results on four grade-school reasoning datasets across three VLM architectures.
On Qwen3-VL-4B-Instruct, our method achieves 55.4\% average accuracy with only 32 latent tokens, compared to 79.3\% accuracy with 108.4 tokens for explicit CoT.
Notably, on simpler tasks such as MultiArith, Render-of-Thought achieves near-parity performance with a 1.8$\times$ reduction in token consumption.
The consistent improvements across all three model architectures validate the generalizability of our approach.

\boldparagraph{Compared with LLM based Latent Reasoning. \\}
Tab.~\ref{tab_compared_with_LLM_methods} compares Render-of-Thought against LLM-based baselines across four grade-school level reasoning datasets.
To ensure fair comparison, all baselines are reproduced using Qwen3-4B-Instruct~\cite{qwen3} as the base model.
Render-of-Thought achieves an average accuracy of 55.4\%, outperforming the best LLM based method, CoLaR-2, by 8.1\%.
While CoLaR-2 yields slightly higher accuracy on GSM8k-Aug, our approach demonstrates superior robustness in out-of-domain generalization. 
We attribute this to the rich semantic representations from the pre-trained visual encoder, which provide more informative supervision signals than the latent spaces learned from scratch in LLM based methods.

\boldparagraph{Performance on High-Difficulty Tasks.}
To assess scalability on more challenging reasoning tasks, we evaluate our method on the MATH dataset.
As detailed in Tab.~\ref{tab_high_difficulty}, we employ three distinct model architectures to demonstrate robustness.
On Qwen3-VL-4B-Instruct, explicit CoT method achieves 55.8\% accuracy but requires an average of 291.5 tokens for the reasoning chain.
In contrast, Render-of-Thought achieves 33.2\% Pass@1 using only 64 latent tokens, surpassing the w/o CoT baseline of 29.4\%.
Furthermore, meaningful improvements over the w/o CoT baseline on the smaller Qwen3-VL-2B-Instruct validate the generalizability of our approach across model scales.

\begin{table*}[t]
\centering
\fontsize{10pt}{11.5pt}\selectfont
% \small
\begin{tabular}{l|cc|cc|cc}
\toprule[0.15em]
& \multicolumn{2}{c|}{Qwen3-VL-2B-Instruct} & \multicolumn{2}{c|}{Qwen3-VL-4B-Instruct} & \multicolumn{2}{c}{LLaVa-V1.6-Mistral-7B} \\
& Pass@1 & \#L & Pass@1 & \#L & Pass@1 & \#L \\ 
\toprule[0.1em]
SFT-w/o CoT & 20.8$_{\pm.21}$ & 0.00$_{\pm.00}$ & 29.4$_{\pm.34}$ & 0.00$_{\pm.00}$ & 11.2$_{\pm.30}$ & 0.00$_{\pm.00}$ \\ 
SFT-CoT & 29.2$_{\pm.29}$ & 324.5$_{\pm2.6}$ & 55.8$_{\pm.36}$ & 291.5$_{\pm1.9}$ & 13.8$_{\pm.33}$ & 200.9$_{\pm2.1}$ \\ 
\toprule[0.1em]
\textbf{RoT (Ours)} & 24.0$_{\pm.22}$ & 64.0$_{\pm.00}$ & 33.2$_{\pm.37}$ & 64.0$_{\pm.00}$ & 12.4$_{\pm.25}$ & 64.0$_{\pm.00}$ \\
\rowcolor[HTML]{E0E0E0}
~~-~w/o Stage I & 15.8$_{\pm.19}$ & 64.0$_{\pm.00}$ & 22.2$_{\pm.34}$ & 64.0$_{\pm.00}$ & 9.40$_{\pm.26}$ & 64.0$_{\pm.00}$ \\
\rowcolor[HTML]{E0E0E0}
~~-~w/o Stage II & 19.2$_{\pm.21}$ & 64.0$_{\pm.00}$ & 26.2$_{\pm.38}$ & 64.0$_{\pm.00}$ & 10.8$_{\pm.28}$ & 64.0$_{\pm.00}$ \\
\bottomrule[0.15em]
% Render-of-Thought         & $14.3_{\pm.25}$~\textcolor{mygreen}{($5.36\%\uparrow$)} & $9.79_{\pm.40}$~\textcolor{mygreen}{($82.8\%\downarrow$)} &       $7.08_{\pm.07}$~\textcolor{mygreen}{($1.80\%\uparrow$)}          &     $16.1_{\pm.14}$~\textcolor{mygreen}{($80.6\%\downarrow$)} &       $7.08_{\pm.07}$~\textcolor{mygreen}{($1.80\%\uparrow$)}          &     $16.1_{\pm.14}$~\textcolor{mygreen}{($80.6\%\downarrow$)}          \\
\end{tabular}
\caption{Experimental results on the challenging MATH dataset across three VLM architectures. Render-of-Thought achieves significant token compression compared to explicit CoT while maintaining competitive accuracy.}
% \caption{Experimental results on the challenging MATH dataset. We evaluate our proposed method \ModelAbbr~on two base models and three settings: -DL denotes using a Deterministic Latent head, -NLL denotes \ModelAbbr~trained with NLL Loss as $\mathcal{L}_\text{latent}$, which is our main method, and -~/w~GRPO denotes the post-trained \ModelAbbr-NLL with GRPO reinforcement learning process. We calculate the performance gain between \ModelAbbr-NLL and \ModelAbbr-NLL-RL to highlight the effectiveness of reinforcement learning. Compression factor $c$ and \#~$\text{L}_{max}$ are set to 2 and 128, respectively.}
\vspace{-6pt}
\label{tab_high_difficulty}
\end{table*}

\begin{figure}
    \centering
    \includegraphics[width=1.0\linewidth]{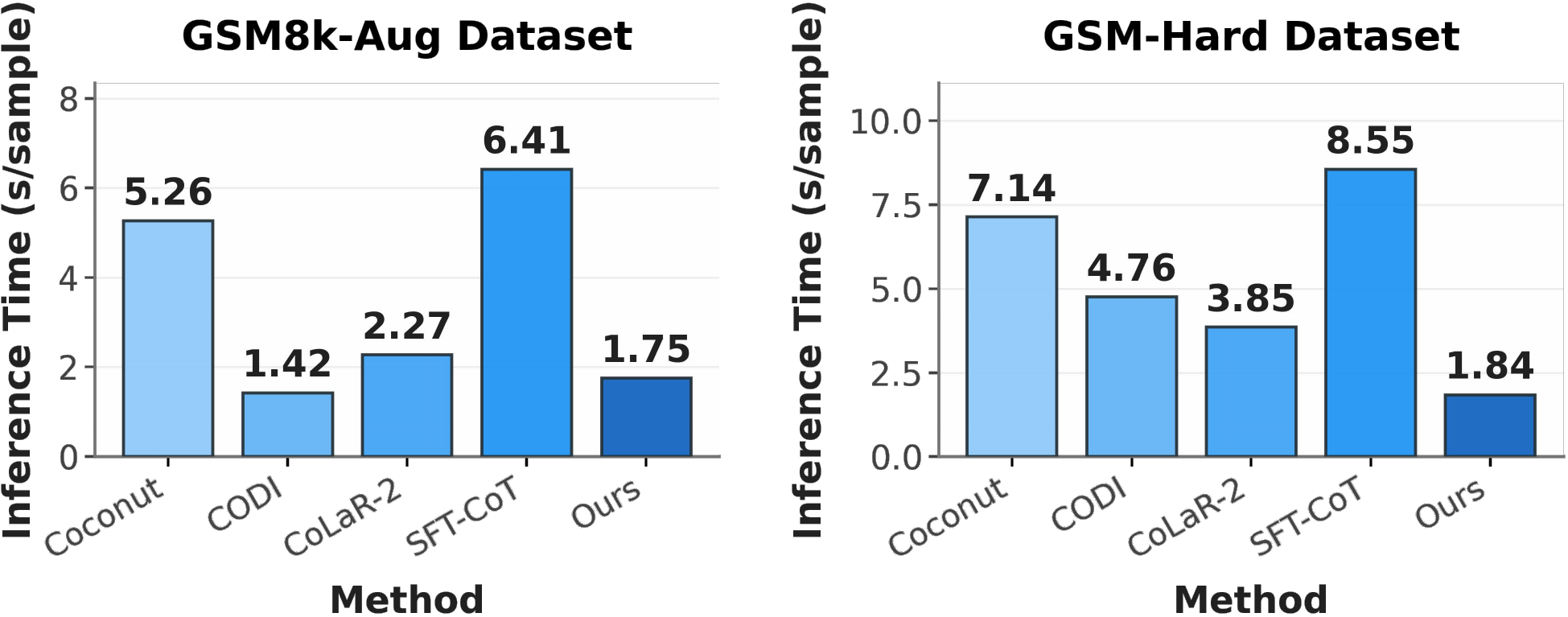}%
    \caption{\textbf{Inference Time Comparison.} We evaluate the average inference time (seconds per sample) on GSM8k-Aug and GSM-Hard datasets using Qwen3-4B-Instruct/Qwen3-VL-4B-Instruct.}
    \label{fig:infer_time}
    \vspace{-6pt}
 \end{figure}

\boldparagraph{Inference Time Comparison.}
We further analyze computational efficiency by comparing the average per-sample inference time on the GSM8k-Aug and GSM-Hard datasets.
To ensure fair comparison, all experiments are conducted on a single NVIDIA H20 GPU with a batch size of 1.
As shown in Fig.~\ref{fig:infer_time}, Render-of-Thought demonstrates significant efficiency gains over explicit CoT.
On the challenging GSM-Hard dataset, inference time decreases notably from 8.55s to 1.84s.
This substantial reduction in latency stems from compressing lengthy textual thoughts into compact sequences of visual latent embeddings. 
Moreover, our method surpasses several latent reasoning baselines in speed, validating the efficiency of the visual latent space.

\subsection{Ablation Study \& Analysis}
\label{ablation study}
\boldparagraph{Effectiveness of Two-Stage Training.}
To assess the contribution of our progressive training strategy, we ablate each stage independently.
Results in Tab.~\ref{tab_compared_with_LLM_methods} and Tab.~\ref{tab_high_difficulty} confirm that both stages are essential for optimal reasoning performance.
Removing Stage I causes accuracy on GSM8k-Aug to drop from 37.8\% to 24.8\%, indicating that visual alignment is vital for structuring the latent space and preventing representation collapse during complex tasks.
Similarly, excluding Stage II leads to a significant performance decline because the model struggles to navigate the continuous latent space toward the final answer.
This necessity is further evidenced on the MATH benchmark where performance falls from 33.2\% to 26.2\% in the absence of Stage II.
These findings demonstrate that our two-stage framework establishes a robust foundation for compressing verbose textual chains into efficient visual latent representations.

\begin{figure}
    \centering
    \includegraphics[width=0.9\linewidth]{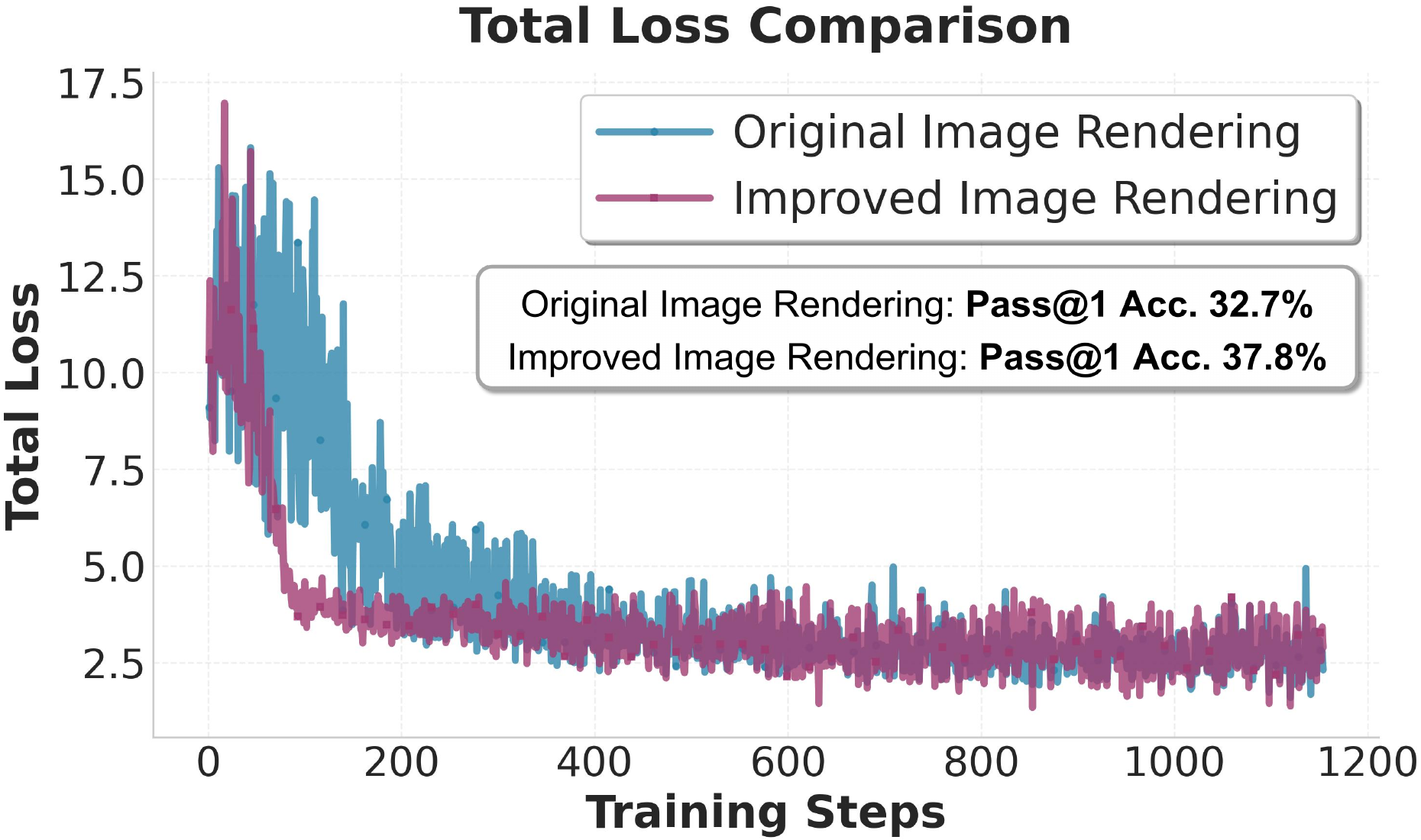}%
    \caption{\textbf{Impact of Rendering Strategies on Training Convergence.} The improved \textbf{\textcolor[RGB]{147,79,128}{single-line rendering}} demonstrates superior stability and speed compared to the \textbf{\textcolor[RGB]{87,157,187}{fixed-size square}} approach.}
    \label{fig:loss_curve}
    \vspace{-6pt}
 \end{figure}

\boldparagraph{Impact of Visual Rendering Configurations.}
The design of visual rendering configurations significantly influences the effectiveness of latent reasoning. We compare two rendering paradigms on GSM8k-Aug dataset: the original approach using fixed-size square images (1024 px × 1024 px) with multi-line text wrapping, and our improved single-line rendering with dynamic width and fixed 32 px height.
As illustrated in Fig.~\ref{fig:loss_curve}, the single-line configuration demonstrates superior convergence. 
Quantitatively, the single-line rendering achieves 37.8\% Pass@1 accuracy on GSM8k-Aug, outperforming the fixed-size square baseline.
This improvement stems from several key design choices. 
First, dynamic width eliminates large blank regions that would otherwise produce meaningless embeddings after visual encoding, preventing the model from learning spurious patterns.
Second, preserving complete text content without truncation ensures no information loss during the rendering process.
Finally, the single-line format is more compatible with sequential modeling paradigms, as it naturally represents reasoning steps as a continuous visual sequence rather than discrete multi-line blocks.
More ablation study results regarding the visual rendering configuration are provided in Appendix Sec.~\ref{ablation study on visual rendering configurations}.

\begin{figure*}
   \centering
   \includegraphics[width=1.0\linewidth]{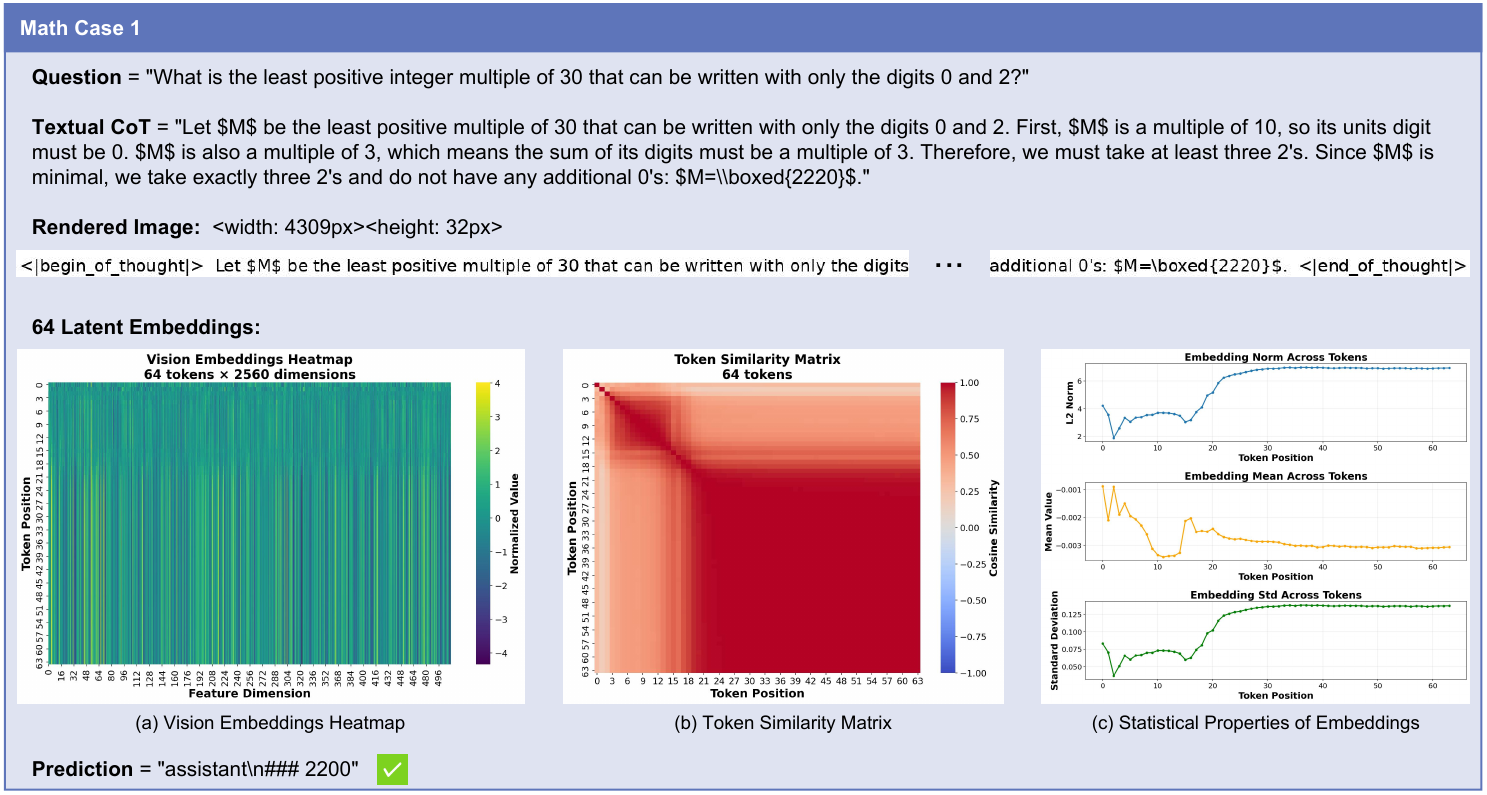}
   \caption{\textbf{Characterizations of Latent Visual Tokens}. We present a case from the MATH dataset. The generated latent embeddings are analyzed via (a) Vision Embeddings Heatmap, (b) Token Similarity Matrix, and (c) Statistical Properties, demonstrating the structured semantic encoding within the continuous visual latent space.}
   \vspace{-8pt}
   \label{fig:vis case}
\end{figure*}

\begin{table}[h]
\centering
\fontsize{10pt}{11pt}\selectfont
\begin{tabular}{l|c|c}
\toprule[0.15em]
\multirow{2}*{Decoding Strategies} & \multicolumn{1}{c|}{GSM8k-Aug} & \multicolumn{1}{c}{MATH} \\
& Pass@1 & Pass@1 \\ 
\toprule[0.1em]
Special Tokens & 3.87 & 2.20 \\
\toprule[0.1em]
Fixed Token Budgets & - & - \\
~~~-~8 tokens & 1.89 & 0.80 \\
~~~-~16 tokens & 11.4 & 8.40 \\
~~~-~32 tokens & \best{37.8} & 30.4 \\
~~~-~64 tokens & 36.2 & \best{33.2} \\
~~~-~128 tokens & 34.6 & 33.0 \\
~~~-~256 tokens & 32.1 & 31.2 \\
\bottomrule[0.15em]
\end{tabular}
\caption{Comparison of decoding strategies on GSM8k-Aug and MATH datasets using Qwen3-VL-4B-Instruct. Fixed token budgets consistently outperform dynamic termination via special tokens.}
\label{tab_decoding_strategies}
\vspace{-2pt}
\end{table}

\boldparagraph{Comparison of Inference Decoding Strategies.}
\label{decoding strategies}
We evaluate two decoding strategies on Qwen3-VL-4B-Instruct across GSM8k-Aug and MATH datasets, as detailed in Tab.~\ref{tab_decoding_strategies}.
The dynamic termination strategy using \colorbox[HTML]{E0E0E0}{$<\vert\texttt{img\_end}\vert>$} yields significantly lower performance compared to fixed token budgets.
This performance gap stems from the inherent instability of self-regulated stopping in continuous latent spaces.
When generating latent reasoning embeddings, the hidden states may not consistently produce high-confidence predictions for the termination token, leading to premature or delayed transitions that disrupt the reasoning flow.
In contrast, fixed token budgets provide stable, predictable termination points that align better with the sequential nature of visual latent reasoning.

The optimal token budget varies across datasets, reflecting differences in task complexity and reasoning depth.
On GSM8k-Aug, 32 tokens achieve the best performance, while MATH requires 64 tokens to reach peak accuracy.
We speculate that this discrepancy arises because the MATH dataset is more challenging and necessitates longer reasoning chains. 
An insufficient token budget severely constrains the model's ability to encode complex reasoning trajectories, resulting in significant performance degradation. Conversely, an excessive budget introduces redundancy and potential noise.
These findings demonstrate that task-specific token budget calibration is crucial for balancing reasoning completeness with computational efficiency.

\subsection{Discussion of Latent Visual Tokens.}
\label{visualization}
% As shown in Fig. \ref{fig:vis case}, We provide a qualitative analysis of the latent reasoning process by visualizing embedding heatmaps and similarity matrices. This analysis demonstrates the model's ability to encode structured semantic features and maintain logical, sequential progression within the continuous latent space.
% Specifically, the distinct diagonal pattern in the token similarity matrix confirms that the latent states evolve sequentially rather than stagnating, effectively mirroring the step-by-step progression of explicit CoT. Furthermore, the emergence of block-diagonal structures in complex tasks suggests the model implicitly organizes reasoning into functional phases, such as problem decomposition and solution formulation.
We observed a phenomenon in the latent tokens generated by Render-of-Thought.
As illustrated in Fig.~\ref{fig:vis case}, the output tokens tend to become increasingly homogeneous after a certain position in the sequence. Specifically, the values in the token similarity matrix approach 1.0, the feature activation heatmaps become nearly identical, and the statistical properties of the embeddings tend to stabilize. This suggests that the model effectively encodes the core reasoning logic in the initial phase, after which the latent states enter a saturation plateau. These subsequent high-similarity tokens likely serve to maintain the semantic context required for decoding the final answer, rather than introducing new reasoning steps or feature shifts.
More visualization results are available in Appendix Sec.~\ref{case study}.

\section{Conclusion}
We introduce Render-of-Thought, the first framework to compress Chain-of-Thought reasoning by rendering textual steps into visual latent representations.
By leveraging pre-trained vision encoders as semantic anchors, our method aims to address the analyzability issues of prior latent reasoning approaches.
Our two-stage training strategy effectively bridges the modality gap, enabling plug-and-play implementation within standard VLM architectures.
Extensive experiments demonstrate 3-4$\times$ token compression and significant inference acceleration compared to explicit CoT, while maintaining competitive accuracy across mathematical and logical benchmarks.
This work establishes visual rendering as a viable paradigm for efficient and analyzable latent reasoning.

\section*{Limitations}
% While Render-of-Thought demonstrates promising results, several limitations warrant future investigation.
% % First, optimal latent token budgets vary across tasks (e.g., 32 tokens for GSM8k-Aug vs. 64 for MATH), requiring task-specific calibration.
% First, optimal latent token budgets vary across tasks, requiring task-specific calibration.
% Second, the training process incurs additional overhead from rendering textual CoT into images and processing them through the vision encoder, though this cost is eliminated during inference.
% Finally, dynamic termination via special tokens exhibits instability in continuous latent spaces, necessitating fixed token budgets for reliable performance.
While Render-of-Thought demonstrates promising results, several limitations warrant future investigation. First, our evaluation is currently limited to English-language mathematical and logical reasoning tasks. The method's effectiveness on other reasoning domains, such as commonsense reasoning or causal inference, as well as its applicability to non-English languages, remains unexplored. Future work could extend the evaluation to diverse reasoning benchmarks and multilingual settings to assess broader generalizability.

Second, the optimal latent token budget requires task-specific calibration, as evidenced by the different optimal values for GSM8k-Aug (32 tokens) and MATH (64 tokens). This manual tuning process may not be feasible for novel applications where task complexity is unknown a priori. Potential solutions include developing adaptive token budget mechanisms that dynamically adjust based on problem difficulty or learning task-specific budget predictors from problem characteristics.
Furthermore, Render-of-Thought encounters a phenomenon similar to that described in \cite{lvr}, where dynamic termination via special tokens exhibits instability in continuous latent spaces. We also intend to address this issue in future work.

Finally, the training process incurs more computational overhead from rendering textual CoT into images and processing them through the vision encoder, though this cost is eliminated during inference. Future work could investigate more efficient rendering strategies or explore caching mechanisms to reduce training time for large-scale deployments.

\section*{Ethics Statement}
This work utilizes publicly available datasets for mathematical and logical reasoning, including GSM8k-Aug~\cite{deng2023implicit}, GSM8k~\cite{gsm8k}, GSM-Hard~\cite{gsm8k-hard}, SVAMP~\cite{svamp}, MultiArith~\cite{multiarith}, and MATH~\cite{math-500}. 
These datasets contain grade-school and challenging mathematical problems that do not involve personal information, sensitive content, or potentially harmful material. 
All datasets are used in accordance with their original licenses and intended research purposes.
Regarding data privacy and protection, we conducted a specific assessment to verify whether the data contains information that names or uniquely identifies individual people. Given that the datasets (e.g., GSM8k, MATH) consist of standard mathematical word problems where names are generic and fictional, we determined that the data does not refer to real-world individuals. Consequently, no additional anonymization or de-identification steps were required beyond the standard usage of these public benchmarks.

We employ open-source vision-language models including Qwen3-VL-2B/4B-Instruct~\cite{qwen3vl} and LLaVa-V1.6-Mistral-7B~\cite{llava}, accessed through standard model repositories such as Hugging Face Hub~\cite{transformers}. 
All models are used under their respective licenses, which permit research use. We have reviewed and complied with all terms of use for these models and their associated training data.

Our training pipeline involves rendering textual Chain-of-Thought annotations into images, which are then processed through pre-trained vision encoders. The rendered images contain only mathematical reasoning steps and problem solutions, without any personal data or offensive content. We have manually inspected a sample of rendered images to ensure they do not contain inappropriate material, and our observations confirm that all rendered content consists solely of mathematical expressions and reasoning steps.

% \section*{Acknowledgments}

% Bibliography entries for the entire Anthology, followed by custom entries
%\bibliography{custom,anthology-overleaf-1,anthology-overleaf-2}

% Custom bibliography entries only
\bibliography{custom}

\maketitlesupplementary
\appendix

\begin{strip}
    \vspace{-50pt}
    \subsection*{Content}
    
    This Appendix contains the following parts:
    \begin{itemize}
        \item \textbf{More Implementation Details}. We provide detailed implementation specifications including model hyperparameters, training hyperparameters, and dataset information.
        \item \textbf{Ablation Study on Visual Projection Head}. We conduct ablation studies on the activation function and hidden dimension of the Visual Projection Head, demonstrating the optimal architectural choices.
        \item \textbf{Ablation Study on Visual Rendering Configurations}. We investigate the influence of rendering configurations, identifying the most effective visual configuration.
        \item \textbf{Training Cost}. We present a quantitative analysis of the training overhead, comparing our method with standard SFT-CoT to evaluate the overall computational efficiency.
        \item \textbf{Unified Backbone Comparison}. We evaluate adapted latent reasoning baselines on a unified VLM backbone, validating that the performance improvements stem directly from our proposed method.
        \item \textbf{Generalization to Non-Mathematical Tasks}. We extend our evaluation to logical and scientific reasoning datasets, demonstrating the broader applicability of our visual rendering paradigm beyond mathematics.
        \item \textbf{Discussion on Dynamic Termination Failure}. We analyze the root causes of dynamic termination failure, attributing it to the fundamental challenges inherent in continuous latent reasoning spaces.
        \item \textbf{Case Study}. We visualize the representations of the latent reasoning embeddings, including heatmaps, similarity matrices and statistical properties, to qualitatively analyze the model's reasoning process across different benchmarks.
    \end{itemize}
    \medskip

    \hrulefill
    \vspace*{4pt} % 在分割线后增加一点小间距
\end{strip}

\section{More Implementation Details}
\label{more implementation details}
\boldparagraph{Model hyperparameters.}
In our experiments, we employ frozen Qwen3-VL-2B/4B-Instruct, LLaVa-V1.6-Mistral-7B, and Qwen3-4B-Instruct as the LLM backbones, utilizing LoRA modules for fine-tuning.
Across all experiments, the LoRA modules are configured with $\alpha=32$, $r=16$, and a dropout rate of 0.05. 
Our method introduces a Visual Projection Head, implemented as a two-layer MLP based on SwiGLU, with the hidden layer dimension set to $d=4096$.

\boldparagraph{Training hyperparameters.}
We utilize the AdamW optimizer with a weight decay of 1e-2 for all experiments. 
The learning rate is set to 2e-5 for both training stages. 
For each training stage, we conduct experiments on two NVIDIA H20 GPUs with DeepSpeed~\cite{deepspeed} configured to Stage 2, using a total batch size of 16.
During Stage I training, the weight $\lambda$ of the alignment loss $\mathcal{L}_{align}$ is set to 10.0.
To ensure reproducibility, we fix the random seeds for all libraries (Python, CUDA, PyTorch, and NumPy) to 0 during the training process.

Additionally, Render-of-Thought involves the special tokens \colorbox[HTML]{E0E0E0}{$<\vert\texttt{img\_begin}\vert>$} and \colorbox[HTML]{E0E0E0}{$<\vert\texttt{img\_end}\vert>$} during training.
For each special token, we first generate a random vector, normalize it to a unit vector, scale it to a norm of $\sqrt{h_d}$, and finally write it into the corresponding position in the embedding table, where $h_d$ represents the hidden dimension of the LLM Backbone.
The use of the $\sqrt{h_d}$ norm is intended to match the typical norm of pre-trained embeddings, ensuring numerical compatibility and training stability.
The random initialization draws reference from LVR~\cite{lvr}, which facilitates the model in distinguishing between the latent reasoning state and the normal semantic state.

\boldparagraph{Dataset information.}
Render-of-Thought is evaluated on five datasets: GSM8K-Aug, GSM8K-Hard, SVAMP, MultiArith, and MATH. Since the original MATH dataset does not provide an official validation set, we follow the protocol of CoLaR~\cite{colar} by randomly shuffling the training set and allocating 10\% of the samples for validation.

\section{Ablation Study on Visual Projection Head}
\label{ablation projection head}
The Visual Projection Head bridges linguistic and visual modalities by mapping LLM hidden states to the visual embedding space. 
To optimize its architecture, we conduct ablation studies on the activation function and hidden dimension. Regarding activation functions, we compare ReLU, GELU\cite{gelu}, and SwiGLU\cite{glu}. 
As shown in Tab.~\ref{tab_projection_head}, SwiGLU consistently outperforms the others. 
We attribute this to its gated mechanism, which enhances feature expressiveness and gradient flow during alignment. 

For the hidden dimension, we find that the default setting of 4096 offers an optimal balance between capacity and efficiency. 
Reducing the dimension to 2048 leads to noticeable performance degradation, particularly on the challenging MATH dataset, underscoring the necessity of sufficient capacity to capture complex reasoning patterns.

\begin{table}[t]
\centering
\fontsize{10pt}{12pt}\selectfont
\begin{tabular}{l|cc}
\toprule[0.15em]
\multirow{2}*{Configuration} & GSM8k-Aug & MATH \\
& Pass@1 & Pass@1 \\ 
\toprule[0.1em]
\multicolumn{3}{c}{\textbf{\second{\textit{Activation Function (Hidden Dim = 4096)}}}} \\
\noalign{\vspace{2pt}}
ReLU & 33.2 & 28.6 \\
GELU & 35.1 & 30.8 \\
SwiGLU & \best{37.8} & \best{33.2} \\
\toprule[0.1em]
\multicolumn{3}{c}{\textbf{\second{\textit{Hidden Dimension (Activation = SwiGLU)}}}} \\
\noalign{\vspace{2pt}}
2048 & 34.5 & 30.1 \\
4096 & \best{37.8} & \best{33.2} \\
\bottomrule[0.15em]
\end{tabular}
\caption{Ablation study on Visual Projection Head configurations using Qwen3-VL-4B-Instruct.}
\label{tab_projection_head}
\end{table}

\section{Ablation Study on Visual Rendering Configurations}
\label{ablation study on visual rendering configurations}
To investigate how rendering hyperparameters influence the semantic encoding capability of the vision encoder, we conduct ablation studies on image height, font size, and padding.
As detailed in Tab.~\ref{tab_rendering_config}, the configuration of 32 px height, 20 px font size, and 4 px padding achieves optimal accuracy.
We observe that image height is a critical factor, reducing it to 16 px results in significant performance degradation. This is likely because insufficient vertical resolution blurs character details, impairing the vision encoder's ability to extract precise textual semantics. Increasing the height to 64 px does not consistently improve performance, suggesting that 32 px generally provide adequate resolution for character legibility without introducing excessive background noise.
Regarding font size, 20 px offers the best performance. Deviating from this optimal size negatively impacts the results, potentially by distorting character stroke features or altering the spatial density to which the pre-trained encoder is adapted.
Finally, appropriate padding (4 px) proves necessary to avoid boundary artifacts, ensuring that character features are fully preserved within the visual receptive field.

\begin{table}[t]
\centering
\fontsize{10pt}{12pt}\selectfont
\begin{tabular}{l|cc}
\toprule[0.15em]
\multirow{2}*{Configuration} & GSM8k-Aug & MATH \\
& Pass@1 & Pass@1 \\ 
\toprule[0.1em]
\multicolumn{3}{c}{\textbf{\second{\textit{Image Height (Font Size = 20, Padding = 4)}}}} \\
\noalign{\vspace{2pt}}
16  px & 34.2 & 29.8 \\
32  px & \best{37.8} & \best{33.2} \\
64  px & 37.1 & 33.5 \\
\toprule[0.1em]
\multicolumn{3}{c}{\textbf{\second{\textit{Font Size (Height = 32, Padding = 4)}}}} \\
\noalign{\vspace{2pt}}
16  px & 35.6 & 31.4 \\
20  px & \best{37.8} & \best{33.2} \\
24  px & 36.9 & 32.7 \\
\toprule[0.1em]
\multicolumn{3}{c}{\textbf{\second{\textit{Padding (Height = 32, Font Size = 20)}}}} \\
\noalign{\vspace{2pt}}
0  px & 37.2 & 32.9 \\
4  px & \best{37.8} & \best{33.2} \\
8  px & 37.5 & 33.0 \\
\bottomrule[0.15em]
\end{tabular}
\caption{Ablation study on visual rendering configurations using Qwen3-VL-4B-Instruct.}
\label{tab_rendering_config}
% \vspace{-8pt}
\end{table}

\section{Training Cost}
We present a quantitative analysis of the training overhead. 
Tab.~\ref{tab_training_time} illustrates the comparison of training times between our method and the standard SFT-CoT on the GSM8k-Aug dataset. 
In our experiments, the training configurations for RoT and the standard SFT-CoT were kept consistent. 
Specifically, the batch size was set to 16, and all experiments were conducted on a single NVIDIA H20 GPU. 
The SFT-CoT model was trained for a total of 3 epochs, whereas RoT underwent 1 epoch of training in Stage I and 2 epochs in Stage II.

\begin{table}[t]
\centering
\fontsize{9pt}{11pt}\selectfont
\begin{tabular}{l|ccc}
\toprule[0.15em]
Method & Stage I & Stage II & Total \\ 
\toprule[0.1em]
SFT-CoT & - & - & 36.9h \\
\textbf{RoT (Ours)} & 19.0h & 37.1h & 56.1h \\		
\bottomrule[0.15em]
\end{tabular}
\caption{Training time comparison on GSM8k-Aug.}
\label{tab_training_time}
% \vspace{-8pt}
\end{table}

\begin{table*}
\centering
\setlength{\tabcolsep}{10pt}
\fontsize{10pt}{12pt}\selectfont
\begin{tabular}{l|c|c|c|c|c}
\toprule[0.15em]
\multirow{2}*{Method} & GSM8k-Aug & GSM-Hard & SVAMP & MultiArith & Average \\
& Pass@1 & Pass@1 & Pass@1 & Pass@1 & Pass@1 \\ 
\toprule[0.1em]
\multicolumn{6}{c}{\textbf{\second{\textit{VLM Based: Qwen3-VL-4B-Instruct}}}} \\
\noalign{\vspace{2pt}}
Coconut & 16.1 & 5.15 & 42.0 & 58.9 & 30.5 \\
CODI & 7.05 & 1.97 & 10.7 & 17.8 & 9.38 \\
CoLaR-2 & \best{39.2} & \secondother{8.18} & \secondother{52.0} & \secondother{81.1} & \secondother{45.3} \\
\textbf{RoT (Ours)} & \secondother{37.8} & \best{14.1} & \best{72.7} & \best{97.2} & \best{55.4} \\		
\bottomrule[0.15em]
\end{tabular}
\caption{Fair Comparison on Same VLM Backbone.}
\label{tab_unified_comparision}
\end{table*}

Due to the processes involved in online rendering and the forward propagation of the visual encoder, training RoT is inherently more time-consuming than SFT-CoT. 
The total actual training duration is approximately 1.5$\times$ that of the text-only SFT-CoT. 
However, this represents a one-time training cost. The inference acceleration achieved during the deployment phase (3-4$\times$ speedup) significantly offsets this overhead. 
Furthermore, adopting an offline rendering strategy could further mitigate the training time costs.

\section{Unified Backbone Comparison}
We provide new comparison results in Tab.~\ref{tab_unified_comparision}, applying latent reasoning baselines to Qwen3-VL-4B-Instruct.
The results show that RoT still outperforms adapted baselines, validating that improvements stem from our method rather than VLM pre-training alone.

\section{Generalization to Non-Mathematical Tasks}
We extended our evaluation to logical and scientific reasoning tasks to demonstrate broader applicability beyond mathematics. 
We further evaluated RoT on the GPQA~\cite{gpqa} and ProsQA~\cite{coconut} datasets. 
The implementation details remain consistent with those described in the original paper. The results, reported as Pass@1 accuracy, are presented in Tab.~\ref{tab_extended_exp}.

\begin{table}[t]
\centering
\setlength{\tabcolsep}{8pt}
\fontsize{10pt}{12pt}\selectfont
\begin{tabular}{l|cc}
\toprule[0.15em]
\multirow{2}*{Method} & GPQA & ProsQA \\
& Pass@1 & Pass@1 \\ 
\toprule[0.1em]
\multicolumn{3}{c}{\textbf{\second{\textit{VLM Based: Qwen3-VL-4B-Instruct}}}} \\
\noalign{\vspace{2pt}}
Coconut & 50.2 & \secondother{99.6} \\
CODI & 51.4 & 98.6 \\
CoLaR-2 & \secondother{59.4} & 99.4 \\
\textbf{RoT (Ours)} & \best{60.5} & \best{100.0} \\
\bottomrule[0.15em]
\end{tabular}
\caption{Results on GPQA and ProsQA Dataset.}
\label{tab_extended_exp}
\end{table}

The experimental results indicate that RoT achieves superior performance compared to previous latent space reasoning approaches on both non-mathematical reasoning datasets. 
Specifically, RoT attained Pass@1 accuracies of 60.5 and 100.0 on GPQA and ProsQA, respectively. 
These findings demonstrate that the visual rendering paradigm can effectively generalize to logical reasoning tasks.

\section{Discussion on Dynamic Termination Failure}
This failure is not a flaw of our method, but rather a fundamental challenge inherent to continuous visual latent reasoning spaces, as empirically demonstrated in LVR~\cite{lvr}. 
The root cause lies in the nature of continuous embeddings, unlike discrete text tokens where termination tokens can be learned as distinct symbols, continuous latent representations lack clear decision boundaries. 
When generating latent reasoning embeddings, the hidden states may not consistently produce high-confidence predictions for the termination token, leading to premature or delayed transitions that disrupt the reasoning flow. 
This instability stems from the smooth, continuous nature of the embedding manifold, where the model must navigate from reasoning states to termination states without discrete landmarks.

Notably, a phenomenon similar to that in RoT was also observed in LVR, which similarly adopted a fixed-budget decoding strategy.
We are committed to addressing this limitation in future work.

\section{Case Study}
\label{case study}
In this section, we present a qualitative analysis of the Render-of-Thought framework by visualizing the latent reasoning process across various benchmarks. 
To provide deeper insights into the internal representations, we visualize three key metrics for the generated latent tokens including vision embeddings heatmaps, token similarity matrices, and statistical properties of the embeddings. 
These visualizations allow us to trace the semantic progression of the model within the continuous latent space.

\begin{figure*}
   \centering
   \includegraphics[width=0.95\linewidth]{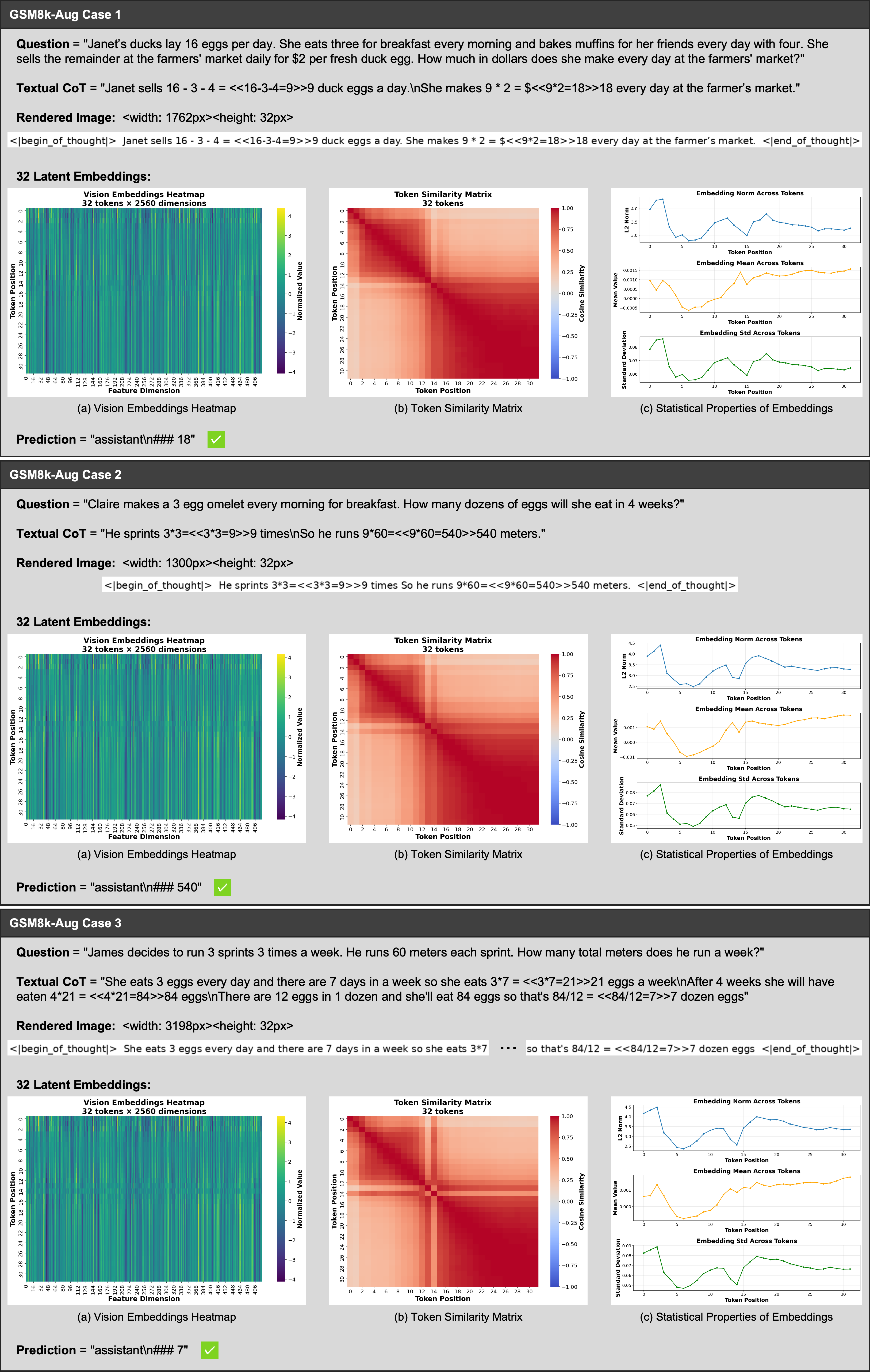}
   \caption{Visualization of reasoning on GSM8k-Aug dataset}
   %\vspace{-6pt}
   \label{fig:gsm8k-aug case}
\end{figure*}

We first examine successful reasoning examples on the GSM8k-Aug dataset as illustrated in Fig.~\ref{fig:gsm8k-aug case}. 
In these instances, the model compresses the reasoning path into a fixed budget of 32 latent embeddings. 
The token similarity matrices exhibit a distinct diagonal pattern with local coherence, suggesting that the model maintains a sequential train of thought where adjacent tokens are semantically related but distinct enough to carry new information. 
Furthermore, the heatmaps display sparse and structured activation patterns, indicating that the model effectively encodes specific semantic features from the visual supervision into the latent space. 
The successful decoding of the final answers demonstrates that 32 latent embeddings are sufficient to capture the reasoning logic for standard grade-school math problems.

\begin{figure*}
   \centering
   \includegraphics[width=0.95\linewidth]{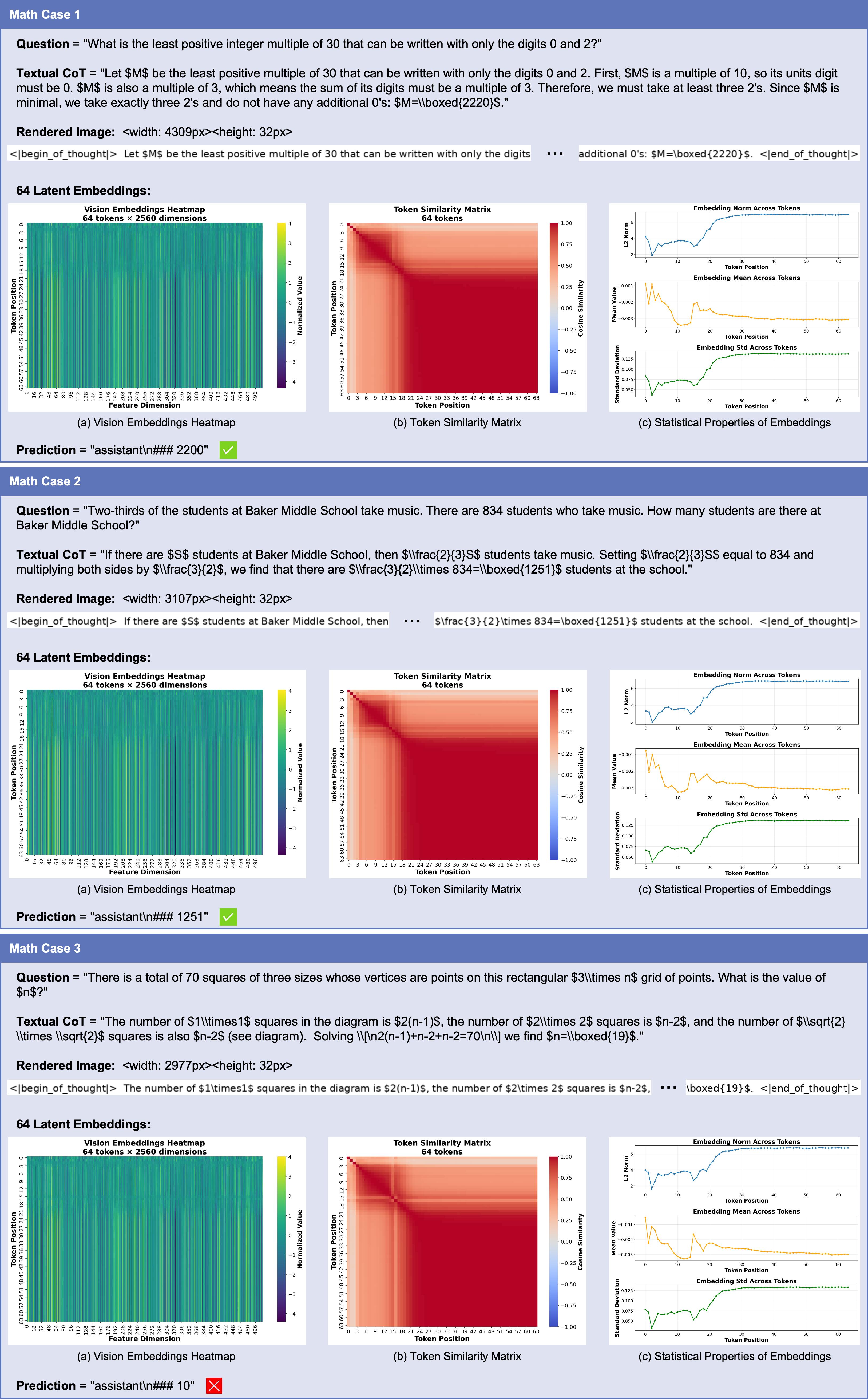}
   \caption{Visualization of reasoning on the challenging MATH dataset.}
   %\vspace{-6pt}
   \label{fig:math case}
\end{figure*}

Fig.~\ref{fig:math case} extends our analysis to the more challenging MATH dataset which involves complex symbols and longer reasoning chains requiring a 64-token budget. As seen in the first two cases, the rendered images contain complex LaTeX expressions that the model successfully aligns with its latent states. The similarity matrices here show a block-diagonal structure that potentially corresponds to different stages of solving the problem, such as understanding the problem and formulating equations.

\begin{figure*}
   \centering
   \includegraphics[width=0.95\linewidth]{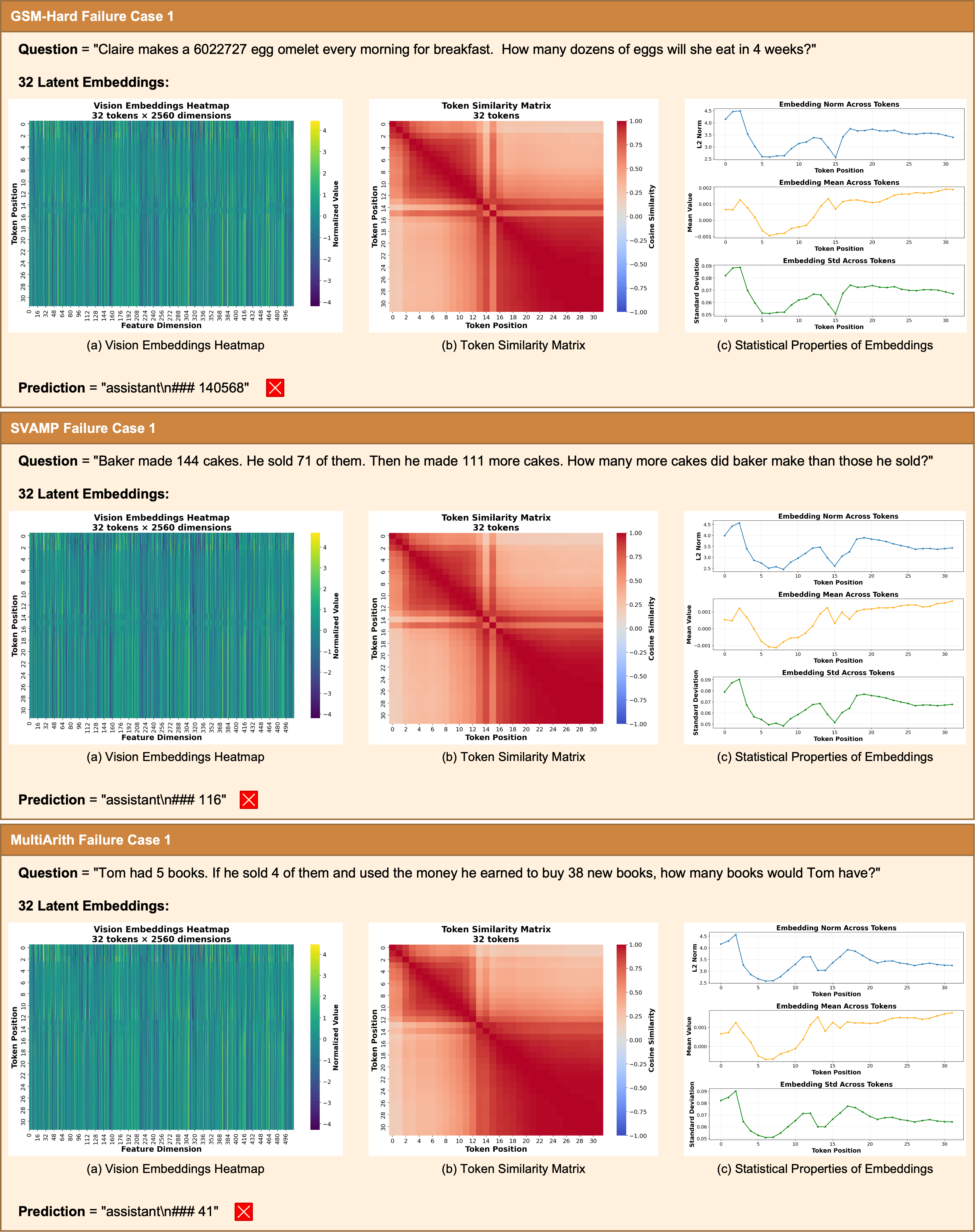}
   \caption{Failure case analysis across out-of-distribution datasets.}
   %\vspace{-6pt}
   \label{fig:failure case}
\end{figure*}

Finally, Fig.~\ref{fig:failure case} highlights failure cases across Out-of-Distribution datasets including GSM-Hard, SVAMP, and MultiArith. 
A common pattern observed in these failure cases is the presence of large and high-similarity blocks in the similarity matrices.
A common pattern observed in these failure cases is the presence of large and highly similar blocks within the similarity matrices. Unlike successful cases, failure cases typically display relatively disordered similarity patterns, implying that the model generates repetitive or indistinguishable latent tokens that fail to effectively advance the reasoning process. Additionally, we observe that failure cases tend to exhibit relatively larger variance. We attribute this to the model's inability to maintain high-confidence representations when encountering unfamiliar problem structures, ultimately leading to incorrect decoding.

\end{document}